\documentclass{article}

\PassOptionsToPackage{numbers, sort}{natbib}

\usepackage[final]{neurips_data_2022}

\usepackage{amsmath,amssymb} %
\usepackage{graphicx}
\usepackage[utf8]{inputenc} %
\usepackage[T1]{fontenc}    %
\usepackage{hyperref}       %
\usepackage{url}            %
\usepackage{booktabs}       %
\usepackage{amsfonts}       %
\usepackage{nicefrac}       %
\usepackage{microtype}      %
\usepackage{xcolor}         %
\usepackage{pifont}%
\usepackage{multirow}

\newcommand{\cmark}{\ding{51}}
\newcommand{\xmark}{\ding{55}}

\newcommand{\changed}[1]{#1} %
\newcommand{\changedagain}[1]{#1} %
\newcommand{\datasetname}{TAP-Vid}

\DeclareMathOperator*{\argmin}{arg\,min}

\usepackage{dsfont} 

\usepackage[frozencache,cachedir=.]{minted}

\usepackage{framed}

\definecolor{formalshade}{rgb}{0.95,0.95,0.95}

\usepackage[strict]{changepage}

\usepackage{colortbl}

\colorlet{shade1}{pink!90!yellow}

\title{TAP-Vid: A Benchmark for Tracking Any Point in a Video}

\author{%
   Carl Doersch$^{*}$ \qquad Ankush Gupta$^{*}$ \qquad Larisa Markeeva$^{*}$ \qquad Adrià Recasens$^{*}$ \\
   \bf Lucas Smaira$^{*}$ \qquad Yusuf Aytar$^{*}$ \qquad João Carreira$^{*}$ \qquad Andrew Zisserman$^{*\dagger}$ \qquad Yi Yang$^{*}$ \\
    \vspace{.1em}\\
    $^{*}$DeepMind \qquad $^{\dagger}$VGG, Department of Engineering Science, University of Oxford
}

\begin{document}

\maketitle

\begin{abstract}
Generic motion understanding from video involves not only tracking objects, but also perceiving how their \emph{surfaces} deform and move.  This information is useful to make inferences about 3D shape, physical properties and object interactions.
While the problem of tracking arbitrary physical points on surfaces over longer video clips has received some attention, no dataset or benchmark for evaluation existed, until now.
In this paper, we first formalize the problem, naming it \textit{tracking any point (TAP)}. We introduce a companion benchmark, \textit{\datasetname}, which is composed of both real-world videos with accurate human annotations of point tracks, and synthetic videos with perfect ground-truth point tracks.  
Central to the construction of our benchmark is a novel semi-automatic crowdsourced pipeline which uses optical flow estimates to compensate for easier, short-term motion like camera shake, allowing annotators to focus on harder sections of video. We validate our pipeline on synthetic data and propose a simple end-to-end point tracking model \textit{TAP-Net}, showing that it outperforms all prior methods on our benchmark when trained on synthetic data.   Code and data are available at the following URL: \url{https://github.com/deepmind/tapnet}.
\end{abstract}

\section{Introduction}

Motion is essential for scene understanding: perceiving how people and objects move, accelerate, turn, and deform improves understanding physical properties, taking actions, and inferring intentions. It becomes even more crucial for embodied agents (i.e.\ robots), as many tasks require precise spatial control of objects over time. While substantial attention has been given to motion estimation, e.g., through algorithms for correspondence and tracking, this paper addresses one particularly under-studied problem: long-term motion estimation of points on generic physical surfaces in real scenes.

There is a multitude of tracking algorithms which address this problem, but partially and with different weaknesses, illustrated in Figure~\ref{fig:tasks}.  Popular box and segment tracking algorithms provide limited information about deformation and rotation of surfaces.  Optical flow can track any point on any surface, but only over pairs of frames, with limited ability to estimate occlusion. Keypoint matching algorithms common in structure from motion typically detect sparse interest points and are not designed for deformable and weakly-textured objects.  For some of the most common tracking problems (e.g.\ faces, hands, and human poses), researchers thus build trackers for hand-chosen semantic keypoints, which is extremely useful but not scalable to arbitrary objects and surfaces. 

\begin{figure}
\centering
\includegraphics[width=.99\linewidth]{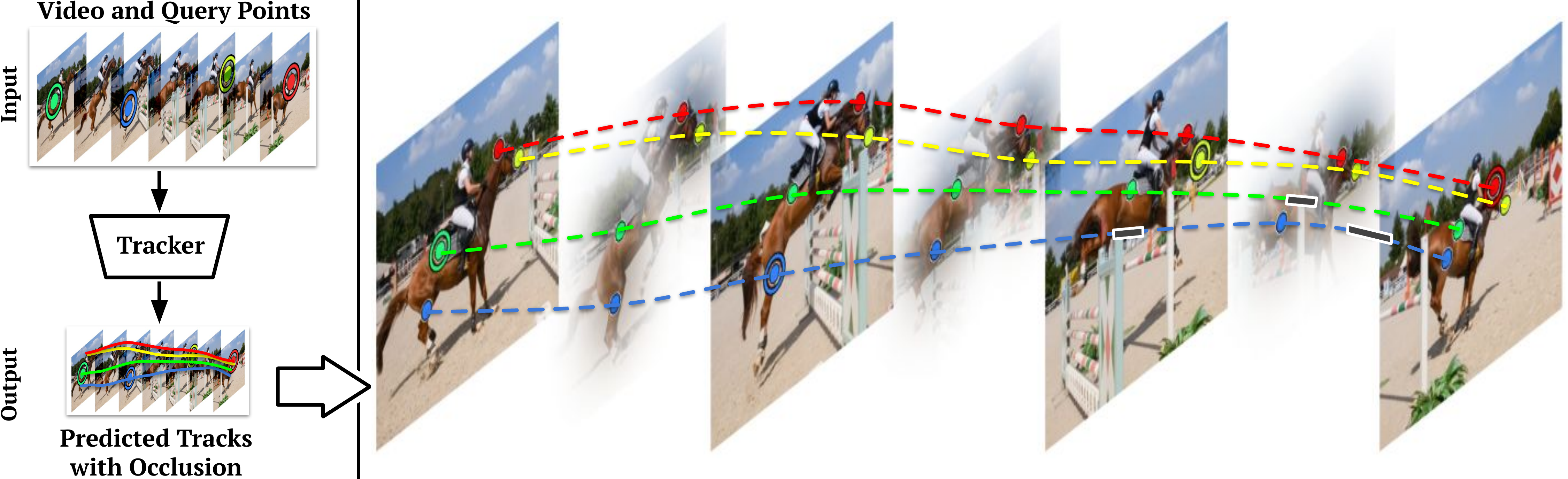}
\caption{{\bf The problem of tracking any point (TAP) in a video.} The input is a video clip (e.g.\ 10s long) and a set of query points ($x,y,t$ in the pixel/frame coordinates; shown with double circles).  The goal is to predict trajectories ($x,y$ pixel coordinates; coloured lines) over the whole video, indicating the same physical point on the same surface, as well as a binary occlusion indicator (black solid segments) indicating frames where it isn't visible.
}
\label{fig:teaser}
\end{figure}

\begin{figure}[t]
\centering
\includegraphics[width=.99\linewidth]{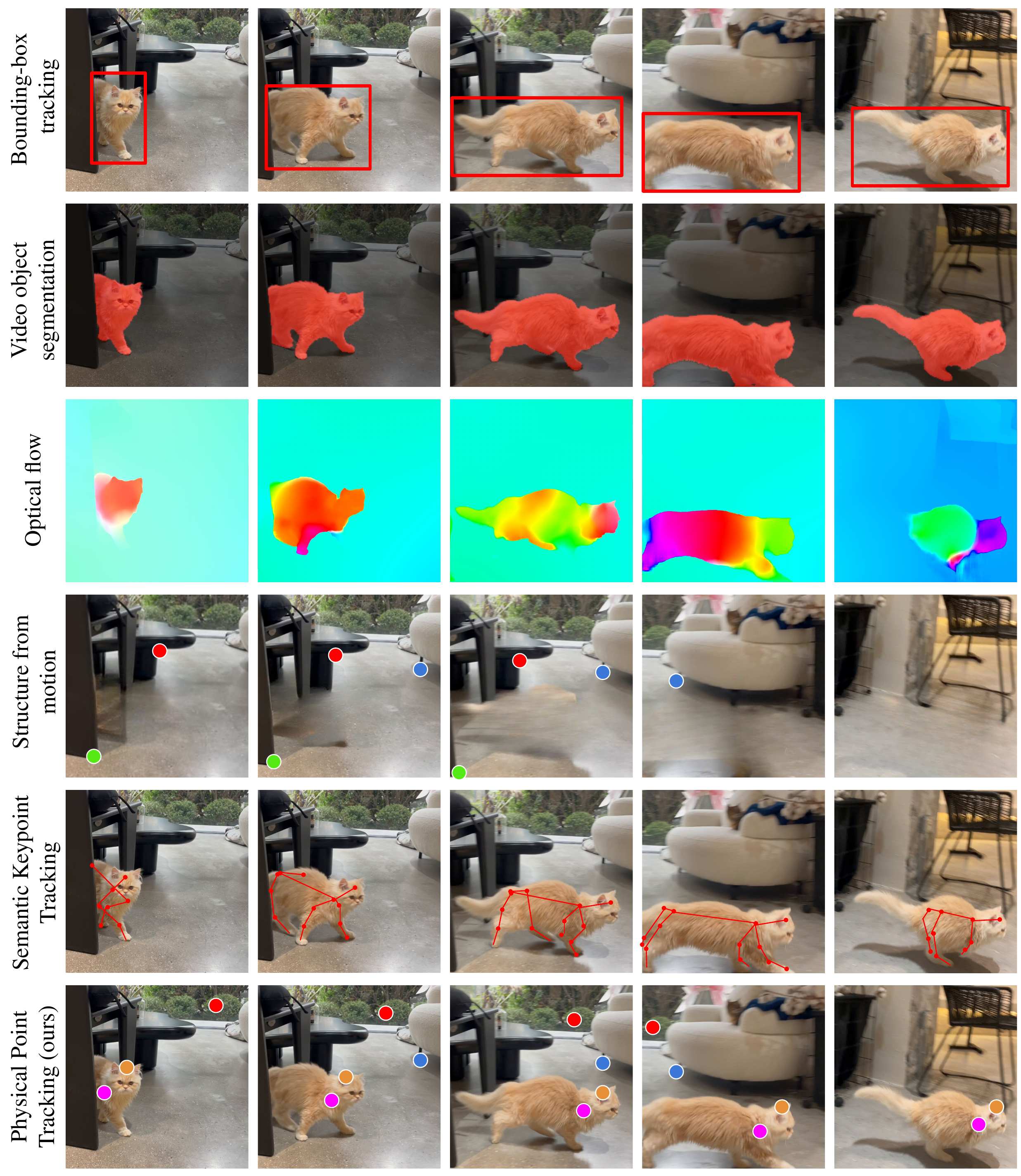}
\caption{{\bf Correspondence tasks in videos.} 
Most prior work on motion understanding has involved tracking (1) bounding boxes or (2) segments, which loses information about rotation and deformation; (3) optical flow which analyzes each frame pair in isolation; (4) structure-from-motion inspired physical keypoints which struggle with deformable objects, or (5) semantic keypoints which are chosen by hand for every object of interest.  Our task, in contrast, is to Track Any Point on physical surfaces, including those on deformable objects, over an entire video.
}

\vspace{-1em}
\label{fig:tasks}
\end{figure}

Our work aims to formalize the problem of long-term physical point tracking directly.  Our formulation, \textit{Tracking Any Point} (TAP) is shown in Figure~\ref{fig:teaser}.  We require only that the target point and surface can be determined unambiguously from a single pixel \textit{query} (i.e.\ we don't deal with transparent objects, liquids, or gasses), and directly estimate how it moves across a video.  \changed{Our task can be seen as an extension of optical flow to longer timeframes with occlusion estimation, or an extension of typical structure-from-motion keypoint matching to typical real-world videos, including e.g. nonrigid or weakly-textured objects and surfaces, also with occlusion estimation.} To train and evaluate, we rely on a mixture of real benchmarks (where humans can directly perceive the correctness of point tracking) and synthetic benchmarks (where even textureless points can be tracked).

While tracking synthetic points is straightforward, obtaining groundtruth labels for arbitrary real-world videos is not.  Thankfully, there's good evidence that humans (and other animals) excel at perceiving whether point tracking is accurate, as it is an example of the ``common fate'' Gestalt principle~\cite{tangemann2021unsupervised,gibson1950perception,todd1990perception,todd1995visual,sturzel2004perceptual}.  
However, \textit{annotating} point tracks in real videos is extremely time-consuming, since both objects and cameras tend to move in complex, non-linear ways; this may explain why this problem has received so little attention in computer vision.  
Therefore, in this work, we first build a pipeline to enable efficient and accurate annotation of point tracks on real videos.
We then use this pipeline to label 1,189 \footnote{The exact number may change due to Youtube wipeout policy.} real YouTube videos from Kinetics~\cite{carreira2017quo} and $30$ DAVIS evaluation videos~\cite{pont20172017} with roughly $25$ points each, using a small pool of skilled annotators and multiple rounds of checking to ensure the accuracy of the annotations. 
Overall we find that on average, roughly 3.3 annotator hours are required to track 30 points through every frame on a 10-second video clip.

The contributions of this paper are threefold.  First, we design and validate an algorithm which assists annotators to more accurately track points.
Second, we build an evaluation dataset with 31,951 (31,301+650) points tracked across 1,219 (1,189 + 30) real videos. 
Third, we explore several baseline algorithms, 
and we compare our point tracking dataset to the closest existing point tracking dataset---JHMDB human keypoint tracking~\cite{jhuang2013towards}---demonstrating that training using our formulation of the problem can increase performance on this far more limited dataset.

\begin{figure}[t]
\centering
\includegraphics[width=.99\linewidth]{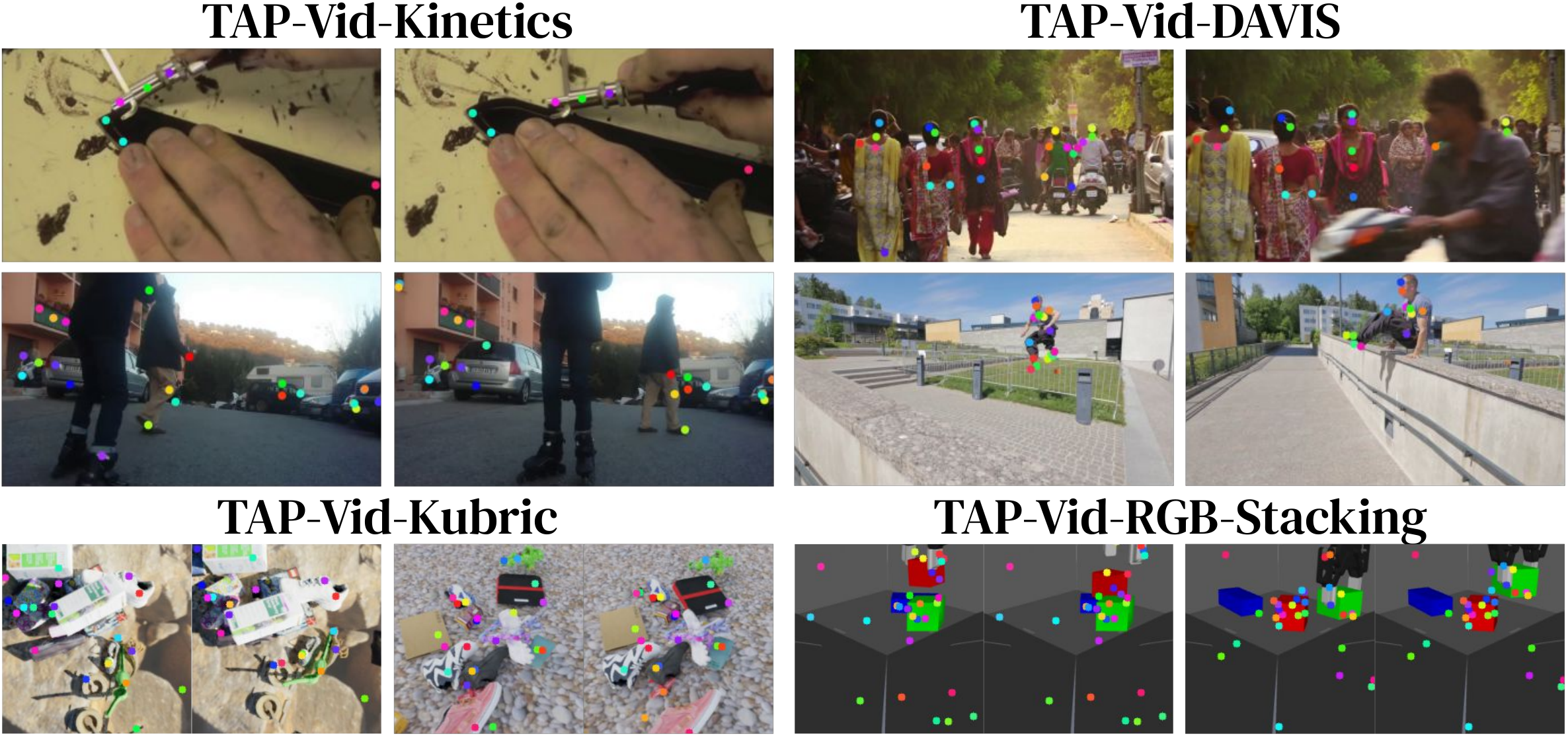}
\caption{{\bf The TAP-Vid point tracking datasets.} Ground-truth point annotations on two random videos from four point tracking datasets we use for evaluation---TAP-Vid-Kinetics and TAP-Vid-DAVIS containing real-world videos with point annotations collected from humans, the synthetic TAP-Vid-Kubric dataset, and TAP-Vid-RGB-Stacking from simulated robotics environment.
}
\label{fig:datasets}
\end{figure}

\section{Related Work}
\label{sec:related}

Though box and segment tracking has been widely benchmarked~\cite{dave2020tao,huang2019got,milan2016mot16,wu2013online,yilmaz2006object,pont20172017,wang2021unidentified,xu2018youtube}, we focus here on methods that track points.  A few early hand-engineered works address long-term tracking of surface points~\cite{wang2013action,sand2008particle,sethi1987finding,lowe1999object,lowe2004distinctive,bay2006surf}, though without learning, these approaches are brittle.
\textbf{Optical Flow} estimates dense motion, but only between image pairs \cite{horn1981determining,lucas1981iterative}; modern approaches~\cite{dosovitskiy2015flownet,ilg2017flownet,ranjan2017optical,sun2018pwc,teed2020raft} train and evaluate on synthetic scenes~\cite{butler2012naturalistic,dosovitskiy2015flownet,mayer2016large,sun2021autoflow}, though depth scanners provide limited real data~\cite{geiger2012we}.
\textbf{Structure-From-Motion (SFM)}~\cite{detone2018superpoint,ono2018lf,jiang2021cotr,manuelli2020keypoints} relies on sparse keypoint matches given image pairs of rigid scenes, and typically ignores occlusion, as it can use geometry to filter errors \cite{hartley2003multiple,schonberger2016structure,torr1999feature,vijayanarasimhan2017sfm}. 
This work is thus often restricted to rigid scenes \cite{zhang2000flexible,marr1979computational}, and is benchmarked accordingly~\cite{dai2017scannet,schops2017multi,balntas2017hpatches,li2018megadepth,zhou2018stereo,li2019learning,li2019learning}.  
\textbf{Semantic Keypoint Tracking} typically chooses a (very small) set of object categories and hand-defines a small number of keypoints to track~\cite{jhuang2013towards,ramanan2005strike,tompson2014real,shrivastava2017learning,xiong2013supervised,sun2014deep}, though some variants track densely for humans~\cite{von2018recovering,densepose}.
\textbf{Keypoint discovery} aims to discover keypoints and correspondence on more general objects, typically with the goal of robotic manipulation~\cite{jakab2018unsupervised,zhang2018unsupervised,jakab2020self,thewlis2017unsupervised,florence2019self}, but these typically train from large datasets depicting a single object.  
Comparison with existing tracking datasets is summarized in~Table~\ref{tab:dataset_comparison}.

\changed{One notable approach which combines several of the above paradigms is a concurrent work PIPs~\cite{harley2022particle}.  This work learns a form of long-term video point tracking inspired by Particle Videos~\cite{sand2008particle}, using a sim2real setup similar to TAP-Net.  However, it struggles to find both training and testing data, relying principally on BADJA~\cite{biggs2018creatures} for evaluation, which contains just 9 videos and tracks joints rather than surface points.}

Our work is also related to "smart" software that aids human annotators, making guesses that annotators may accept or reject, so they don't need to label exhaustively. Polygon RNN~\cite{acuna2018efficient} and UVO segmentation~\cite{wang2021unidentified} accelerated image and video segmentation respectively, and similar systems have been proposed for object tracking in videos~\cite{Kuznetsova_2021_WACV,vondrick2011video,vondrick2013efficiently}.  

\begin{table*}[t]
\setlength{\tabcolsep}{2pt}
\resizebox{\textwidth}{!}{%
\begin{tabular}{llrrcccc}
\hline
\multicolumn{1}{c}{\textbf{Dataset}}    & \multicolumn{1}{c}{\textbf{Type}} & \multicolumn{1}{c}{\textbf{\begin{tabular}[c]{@{}c@{}}\#Videos \\ (\#Images)\end{tabular}}} & \multicolumn{1}{c}{\textbf{\begin{tabular}[c]{@{}c@{}}Duration/\\ Time-scale\end{tabular}}} & \textbf{\begin{tabular}[c]{@{}c@{}}Long-\\ term?\end{tabular}} & \textbf{\begin{tabular}[c]{@{}c@{}}Point \\ Precise?\end{tabular}} & \textbf{\begin{tabular}[c]{@{}c@{}}Class \\ Agnostic?\end{tabular}} & \textbf{\begin{tabular}[c]{@{}c@{}}Non-\\ rigid?\end{tabular}} \\ \hline
KITTI~\cite{geiger2012we}               & Optical flow                      & 156                                                                                       & 400 frame pairs                                                                             & \xmark                                                         & \cmark                                                             & \cmark                                                              & \cmark                                                         \\
COCO-DensePose~\cite{densepose}         & Surface Points                    & (50k)                                                                                     & ---                                                                                         & \xmark                                                         & \cmark                                                             & \xmark                                                              & \cmark                                                         \\
COCO-WholeBody~\cite{jin2020whole}      & Semantic Keypoints                & (200k)                                                                                    & ---                                                                                         & \xmark                                                         & \cmark                                                             & \xmark                                                              & \cmark                                                         \\
DAVIS~\cite{pont20172017}               & Masks (multi-object)              & 150                                                                                       & 25 fps @ 2-5s                                                                               & \cmark                                                         & \xmark                                                             & \cmark                                                              & \cmark                                                         \\
GOT-10k~\cite{huang2019got}             & BBs (single-object)               & 10k                                                                                       & 10 fps @ 15s                                                                                & \cmark                                                         & \xmark                                                             & \cmark                                                              & \cmark                                                         \\
TAO~\cite{dave2020tao}                  & BBs (multi-object)                & 3k                                                                                        & 1 fps @ 37s                                                                                 & \cmark                                                         & \xmark                                                             & \cmark                                                              & \cmark                                                         \\
YouTube-BB~\cite{real2017youtube}       & BBs (single-object)               & 240k                                                                                      & 1 fps @ 20s                                                                                 & \cmark                                                         & \xmark                                                             & \xmark                                                              & \cmark                                                         \\
PoseTrack~\cite{andriluka2018posetrack} & Semantic Keypoints                & 550 (37k)                                                                                 & \begin{tabular}[c]{@{}r@{}}train: 30 fps @ 1s\\ eval: 7 fps @ 3--5s\end{tabular}            & \cmark                                                         & \cmark                                                             & \xmark                                                              & \cmark                                                         \\
300VW~\cite{shen2015first}              & Facial Keypoints                  & 300                                                                                       & 30 fps @1--2 mins                                                                            & \cmark                                                         & \cmark                                                             & \xmark                                                              & \cmark                                                         \\
ScanNet~\cite{dai2017scannet}           & SfM 3D recons.                    & 1500                                                                                      & $\sim$1 min                                                                                 & \cmark                                                         & \cmark                                                             & \cmark                                                              & \xmark                                                         \\
MegaDepth~\cite{li2018megadepth}        & SfM 3D recons.                    & \begin{tabular}[c]{@{}r@{}}200 scenes \\ (130k)\end{tabular}                              & ---                                                                                         & \cmark                                                         & \cmark                                                             & \cmark                                                              & \xmark                                                         \\ \hline
\textbf{TAP-Vid-Kinetics}               & Arbitrary points                  & 1,189                                                                                      & 25 fps @10s                                                                                 & \cmark                                                         & \cmark                                                             & \cmark                                                              & \cmark                                                         \\ \hline
\end{tabular}}
\caption{{\bf Comparison with tracking datasets.} Our dataset includes precise human annotated tracks of arbitrary class-agnostic points over long(-er) duration (10 seconds), unlike the existing datasets.}
\label{tab:dataset_comparison}
\end{table*}

\begin{table*}[t]
\resizebox{\linewidth}{!}{ %
\begin{tabular}{lcccccc}
\toprule
Dataset & \hspace{.5em} $\#$ Videos \hspace{.5em} & \hspace{.5em} Avg. points \hspace{.5em} & \hspace{.5em} $\#$ Frames \hspace{.5em} & \hspace{.5em} Initial resolution \hspace{.5em} & \hspace{.5em} Sim/Real & \hspace{.5em} Eval resolution \hspace{.5em}\\
\midrule
\datasetname-Kinetics & 1,189 & 26.3 & 250 & $\geq$720p & Real & 256x256\\
\datasetname-DAVIS & 30 & 21.7 & 34-104 & $~$1080p & Real & 256x256\\
TAP-Vid-Kubric & 38,325/799 & flexible & 24 & 256x256 & Sim & 256x256\\
\datasetname-RGB-Stacking & 50 & 30 & 250 & 256x256 & Sim & 256x256\\
\bottomrule
\end{tabular}
}
\caption{
{\bf Statistics of our four TAP-Vid datasets}.  Avg. points is average number of annotated points per video; $\#$ Frames is the number of frames per video. Note, the TAP-Vid-Kubric data loader can sample arbitrary points, and therefore there are functionally unlimited points per Kubric video. 
}
\label{tab:point_tracking}
\end{table*}

\section{Dataset Overview}

Figure~\ref{fig:teaser} illustrates our general problem formulation.  Algorithms receive both a video and a set of \emph{query} points, i.e., a list of points $(x,y,t)$: $x,y$ for 2d position and $t$ for time.  For each query point, the algorithm must output 1) a set of positions $(x_t,y_t)$, one per frame, estimating where that point has moved, and 2) a binary value $o_t$ indicating if the point is occluded on each frame.  During occlusions, the $(x_t,y_t)$ output is typically assumed to be meaningless. 
Our benchmark combines both synthetic datasets---with perfect tracks but imperfect realism---and real datasets, where points must be tracked by hand.  
Note that with our real-world data, we principally aim for an \textit{evaluation} benchmark. 
We expect TAP will be used in domains beyond the ones we annotate, e.g., novel robotics problems that have not yet been formulated.
\textit{Transfer} to evaluation datasets from synthetic data like Kubric is more likely to be representative of performance on unseen domains.  \changedagain{This is thus the avenue we pursue in this work: we use Kubric for training the presented models, and hold out the other three datasets exclusively for testing.}
Note that evaluating the ability of an algorithm to track any point does not require us to label \textit{every} point; the points must simply be a sufficiently random sample of all trackable points.  Therefore, given a finite annotation budget, we prioritize diversity, and label a few points on each of a large set of videos.

For real-world evaluation, we annotate videos from the Kinetics-700 validation subset~\cite{carreira2017quo}, a standard large-scale dataset for video understanding with diverse human actions, and the DAVIS validation set~\cite{pont20172017}, a go-to benchmark for evaluating class-agnostic correspondence via segment tracking.
For synthetic data, we turn to Kubric MOVi-E~\cite{kubric} and  RGB-Stacking~\cite{lee2021beyond}, as an example downstream task relevant to roboticists.
For these synthetic datasets, we rely on the simulator to obtain ground-truth point tracks;
implementation details can be found in Appendix~\ref{sec:sim_data}. Table~\ref{tab:point_tracking} lists basic statistics for our TAP-Vid benchmark, and Figure~\ref{fig:datasets} shows example video frames.
Our annotations are released under a Creative Commons Attribution 4.0 License.

\subsection{TAP-Vid Datasets}

\noindent \textbf{\datasetname-Kinetics.} This dataset consists of videos from the Kinetics-700 validation set~\cite{carreira2017quo}, which represent a diverse set of actions relevant to people.  We randomly sample videos which have 720p resolution available, as this aids annotation accuracy.  These videos are unstructured YouTube clips, often with multiple moving objects, as well as moving or shaking cameras and non-ideal lighting conditions.  Kinetics clips are 10 seconds, which we resample at 25 fps, resulting in 250-frame clips. 

\noindent \textbf{\datasetname-DAVIS.} This dataset consists of videos from DAVIS 2017 validation set~\cite{pont20172017}: 30 videos intended to be challenging for segment tracking. We use the same annotation pipeline, but as DAVIS often contains only one salient object, we ask annotators to label up to 5 objects and 5 points per object.
The point tracks were annotated at 1080p resolution to improve the precision. 
Like with \datasetname-Kinetics, the whole dataset is used for evaluation, resized to 256x256.

\noindent \textbf{\datasetname-Kubric.} We use the synthetic MOVi-E dataset introduced in Kubric~\cite{kubric} as both our main source of supervised training data and also for evaluation.  Each video consists of roughly 20 objects dropped into a synthetic scene, with physics from Bullet~\cite{coumans2015bullet} and raytraced rendering from Blender~\cite{blender}.  For training, we follow prior augmentation algorithms: cropping the video with an aspect ratio up to $2:1$ and as little as 30\% of the pixels.  A validation set of 799 videos from the same distribution is held out for testing.

\noindent \textbf{\datasetname-RGB-Stacking.} This synthetic dataset consists of videos recorded from a simulated robotic stacking environment introduced in \cite{lee2021beyond}.
We record 50 episodes in which teleoperators play with geometric shapes and are asked to display diverse and interesting motions. 
We use the \emph{triplet-4} object set and the front-left camera.
We sample 30 points per video from the first frame (20 on moving objects and 10 static objects/background) and track them via the simulator.
The dataset is particularly challenging because objects are textureless and rotationally symmetric with frequent occlusions from the gripper, which are common in robotic manipulation environments.

\section{Real-World Dataset Construction}

\begin{figure}
\centering
\includegraphics[width=\linewidth]{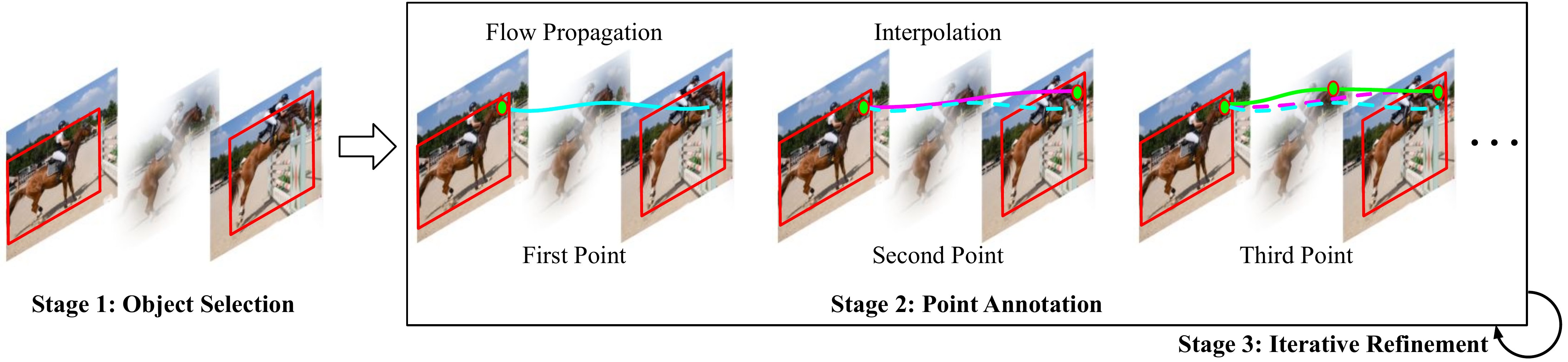}
\caption{{\bf Annotation workflow.} There are 3 stages: (1): object selection with bounding-boxes, (2) point annotation through optical-flow based assistance, and (3) iterative refinement and correction.
}
\label{fig:workflow}
\end{figure}

\begin{figure}[t]
\centering
\includegraphics[width=.99\linewidth]{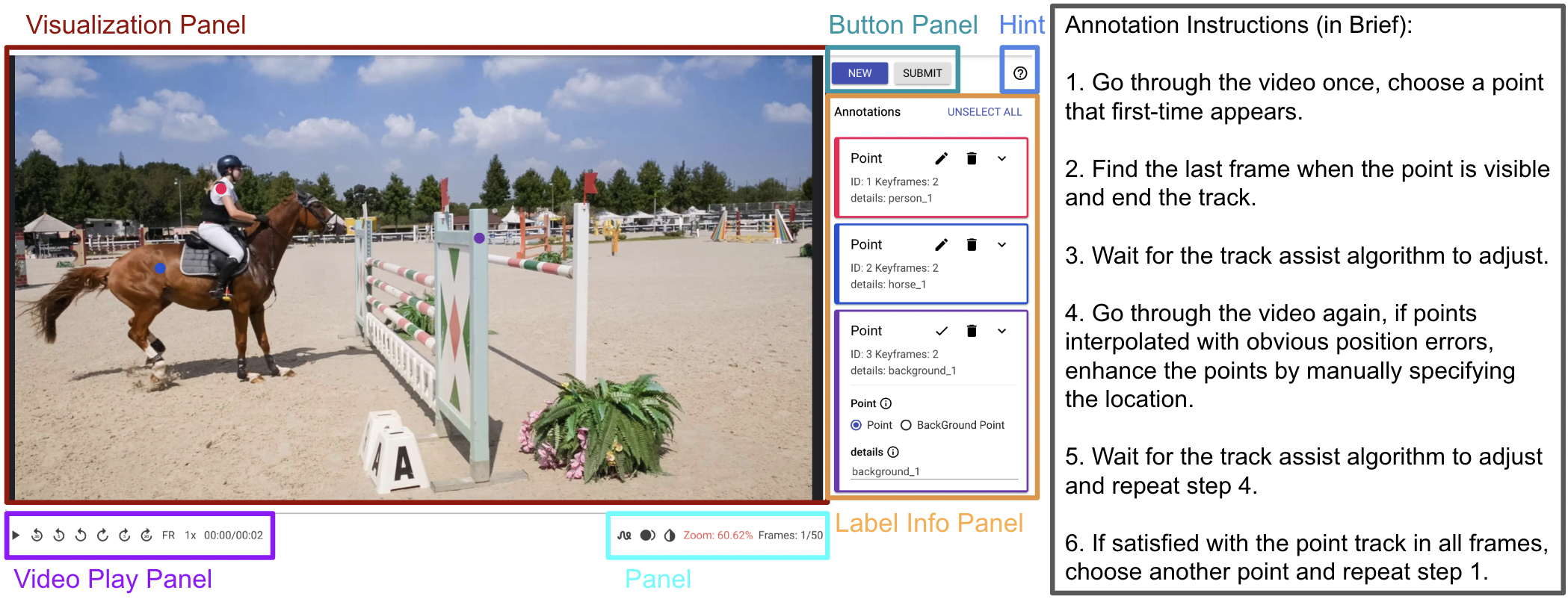}
\caption{{\bf Point annotation interface and instructions.} The interface consists of three components: visualization panel, buttons, and information panels. The instructions consist of six steps which guide annotators to iteratively add points with the help of the track assist algorithm.}
\label{fig:interface}
\end{figure}

Inspired by the Tracking Any Object (TAO) dataset~\cite{dave2020tao}, we aim for generality by allowing annotators to choose any object and any point they consider important, rather than specifying a closed-world set of points to annotate.
Given a video, annotation proceeds in three stages, depicted in Figure~\ref{fig:workflow}.  First, annotators choose objects, especially moving ones, without regard to their difficulty in tracking.  Next, they choose points on each selected object and track them.  
Finally, we have a refinement phase where low-quality annotations are corrected by a different annotator, iterating as many times as needed.  The work was performed by 15 annotators in Google's crowdsourcing pool, paid hourly. Note that, by using a relatively small and stable pool, annotators gain expertise with the tasks and the annotation interface in order to label most effectively.  The full details of our design, including compensation rates, were reviewed by DeepMind's independent ethical review committee. All annotators provided informed consent prior to completing tasks and were reimbursed for their time. 

\textbf{Stage 1: Object Selection}
In the first stage, annotators choose up to $K$ objects ($K=10$ for Kinetics and $K=5$ for DAVIS).
We ask annotators to prioritize objects that are salient and appear for a longer duration. We also prioritize moving objects and those producing sounds.
We define an object as a standalone entity where it exists on its own, e.g.\ car, chair, person, which are solid, avoiding liquids/gasses and transparent objects (see Appendix~\ref{sec:annotation_instructions} for detailed instructions). 
Annotators then draw a box around each object once every 30 frames, using the interface from~\cite{Kuznetsova_2021_WACV} (though we don't use any automated box tracking), and annotators also add a text label per object.

\textbf{Stage 2: Point Annotation}
For every object selected in the first stage, annotators choose a set of $M$ points ($M=3$ for Kinetics and $M=5$ for DAVIS).  Then they must annotate every other frame in the video with the corresponding location (except when occluded; annotators mark frames where points are not visible).  Annotating every other frame makes annotation more accurate, as human eyes are tuned to whether or not a point is attached to a surface from how it moves.  Exhaustively labeling every frame, however, is prohibitive, so we provide a novel \textit{track assist} algorithm, which uses optical flow to convert sparsely-chosen points into dense tracks that follow the estimated motion.
 
Annotators can choose as many or as few points as required to achieve the desired accuracy.

\textbf{Stage 3: Iterative Refinement}
To ensure annotation quality, after initial submission, each annotated point goes to a second annotator, who checks and corrects points to achieve the desired accuracy.
This iterative refinement continues until the last annotator agrees with all previous labels.
On average, a 10s video takes approximately 3.3 hours to finish, often with 4--5 annotators working on the refinement.

\subsection{Annotation Interface}

Figure \ref{fig:interface} shows the point annotation interface based on Kuznetsova~\emph{et~al.}~\cite{Kuznetsova_2021_WACV} presented to annotators in stages 2 and 3. The interface loads and visualizes video in the visualization panel.
Buttons in the video play panel allow annotators to navigate frames during annotation. The information panel provides basic information, e.g., the current and total number of frames. The annotation buttons (\texttt{NEW} and \texttt{SUBMIT}) allow annotators to add new point tracks or submit the labels if finished. The label info panel shows each annotated point track and the associated `tag' string.

Each track begins with an \texttt{ENTER} point, continues with \texttt{MOVE} points, and finishes when the annotator sets it as an \texttt{EXIT} point. Annotators can restart the track after an occlusion by adding another \texttt{ENTER} point and continuing. The cursor is cross shaped which allows annotators to localize more precisely. More details are provided in Appendix~\ref{sec:annotation_instructions}.

\subsection{Track Assist Algorithm}

Given an initial point, modern optical flow algorithms like RAFT~\cite{teed2020raft} can track a point somewhat reliably over a few frames, meaning they have the potential to annotate many frames with a single click.  However, there are two problems: the estimates tend to drift if interpolated across many frames, and they cannot handle occlusions.  
Annotators manually end tracks at occlusions, but to deal with drift, we need an algorithm that allows annotators to \textit{adjust} estimates created via optical flow to compensate for errors that would otherwise accumulate.

We first compute optical flow using RAFT for the entire video.  When the annotator selects a starting point $p_s$ on frame $s$, we use flow to propagate that point from one frame to the next (using bilinear interpolation for fractional pixels) all the way to the last frame.  When the annotator chooses a second point $p_t$ on frame $t$, we find the path which minimizes the squared discrepancy with the optical flow estimated for each frame.  This corresponds to solving the following optimization problem:

\begin{eqnarray}
\argmin_{\rho\in \mathcal{P}_{s:t}}\sum_{i=s}^{t-1}\left\|(\rho_{i+1}-\rho_{i})-\mathcal{F}(p_{i})\right\|^{2} \hspace{.5em} & \mbox{s.t.} & \hspace{.5em} \rho_s = p_s, \rho_t = p_t
\end{eqnarray}

Here, $\mathcal{P}_{s:t}$ is the set of all possible paths from frame $s$ to frame $t$, where each path is described by a list of points.  That is, each element $\rho$ in $\mathcal{P}_{s,t}$ is a list $\{\rho_i, i\in \{s,...,t\}\}$ where $\rho_{i} \in \mathbb{Z}^{2}$.  $\mathcal{F} :  \mathbb{Z}^{2} \rightarrow \mathbb{R}^{2} $ is the optical flow tensor, i.e., it maps image pixels to an optical flow vector.  

While this is a non-convex optimization problem, it can be seen as a
shortest-path problem on a graph where each node corresponds to a pixel in a given frame.  Each pixel in frame $i$ is connected to every pixel in frame $i+1$, with a weight proportional to the squared discrepancy with the optical flow at frame $i$. 
This can be solved efficiently with a form of Dijkstra's algorithm.
In practice, the shortest path is typically found within a few seconds (faster than an annotator can click points) even when points are separated by dozens of frames (a few seconds of video).
Annotators are instructed to check all returned tracks and to `split' unsatisfactory interpolations, by adding more points that become new endpoints for the interpolation.
If two points chosen by the annotators are closer than 5 frames apart, we assume the optical flow estimates are not helpful and interpolate linearly, \changed{and annotators have the option to fall back on linear interpolation at any time}.
\changed{
We find that on average, annotators manually click 10.2 points per each track-segment (i.e., contiguous, un-occluded sub-track). There are about 57 points per segment on average, with a minimum of 2 manually-selected points per segment (the start and end, except for the occasional rare segments which are visible for a single frame).  These results suggest that annotators are actively engaged in removing errors and drift from imperfect automatic propagation.}

\begin{figure}[t]
\centering
    \includegraphics[width=\textwidth,trim=45mm 19mm 50mm 22mm,clip]{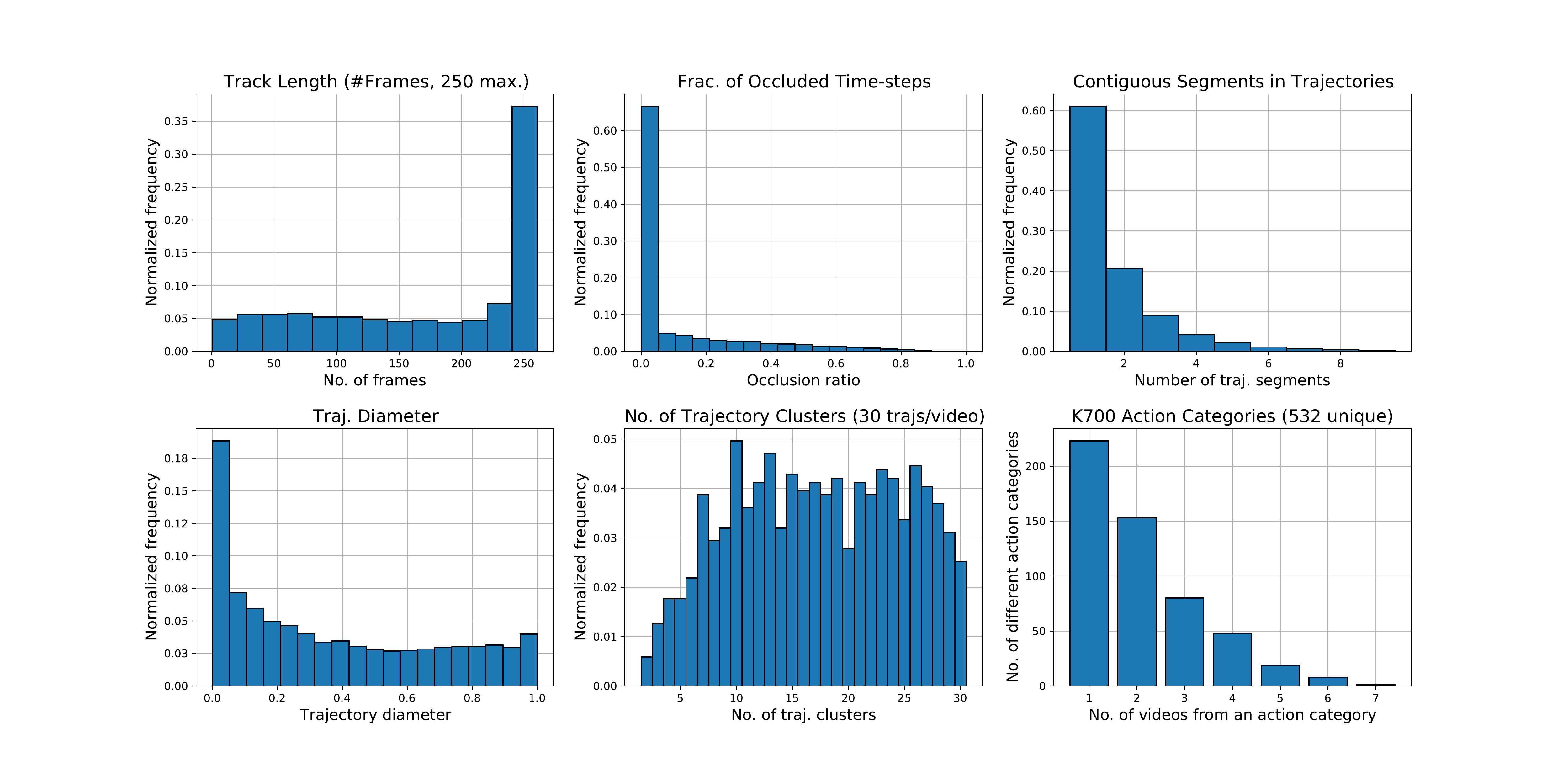}
\caption{{\bf Statistics of trajectories in TAP-Vid-Kinetics.} 
\emph{Diameter}
refers to the maximum distance between the positions of a point over time. \emph{Trajectory Segments} refer to the different number of contiguous sections of point trajectories with breaks due to occlusion. \emph{Trajectory Clusters} are formed from clustering the 30 trajectories per video with the distance between two trajectories being invariant to their relative offset; a large number of clusters indicate diverse point motion.}
\label{fig:stats}
\end{figure}

\subsection{Evaluation and Metrics}    
\label{sec:evaluation_and_metrics}
We aim to provide a simple metric which captures accuracy at both predicting the locations of visible points and predicting occluded points.  However, comparing a binary decision (occlusion) with a regression problem isn't straightforward; therefore, similar to PCK~\cite{yang2012articulated}, we threshold the regression problem so that it becomes binary.  Then, as has been advocated for object box tracking~\cite{luiten2021hota}, we compute a Jaccard-style metric which incorporates both position and occlusion.

Specifically, we adopt three evaluation metrics. (1) \textit{Occlusion Accuracy (OA)} is a simple classification accuracy for the point occlusion prediction on each frame. (2) \textit{$<\delta^x$} evaluates the position accuracy only for frames where the point is visible. Like PCK~\cite{yang2012articulated}, for a given threshold $\delta$, it measures the fraction of points that are within that threshold of their ground truth. While PCK typically scales the threshold relative to a body (e.g., a human), our metric simply assumes images are resized to 256x256 and measures in pixels.  $<\delta^{x}_{avg}$ averages across 5 thresholds: 1,2,4,8, and 16 pixels.  
(3) The final metric, \textit{Jaccard at $\delta$}, evaluates both occlusion and position accuracy.  It's the fraction of `true positives', i.e., points within the threshold of any visible ground truth points, divided by `true positives' plus `false positives' (points that are predicted visible, but the ground truth is either occluded or farther than the threshold) plus `false negatives' (groundtruth visible points that are predicted as occluded or the prediction is farther than the threshold).  \textit{Average Jaccard (AJ)} averages Jaccard across the same thresholds as $<\delta^{x}_{avg}$. 

\changedagain{When evaluating on \datasetname, users are encouraged to train only on \datasetname-Kubric, but validation is a challenge, as \datasetname-Kubric performance may not correlate very well with performance on other datasets due to domain overfitting, especially with larger models.  Therefore, if running a large number of experiments, we encourage users to use \datasetname-DAVIS as a validation dataset and hold out the other datasets purely for testing.  Automated validation (e.g. large-scale architecture search) on the held-out datasets, even \datasetname-DAVIS, is discouraged.}

\section{Dataset Analysis}

We first provide some basic dataset statistics.  Then we validate the accuracy of our point tracking procedure, both qualitatively and manually annotating tracks on synthetic data where they may be compared with ground truth.  Finally, to aid future research, we evaluate several baseline algorithms from the literature.  Finding that they perform relatively poorly, we develop our own somewhat stronger baseline (TAP-Net) using a version of cost volumes inspired from optical flow algorithms.  

\subsection{Point trajectory statistics}
Figure~\ref{fig:stats} summarizes a few key statistics of the point trajectories in our dataset. Most tracks span the entire duration of the clip (250 frames), are not occluded (55\%), and are broken into a small number of segments due to occlusion ($> 70\%$ have $\leq$ 2 segments). 
To quantify the amount of motion a point undergoes, we compute its \emph{trajectory diameter}, i.e., the maximum distance between any two positions of the point over the entire trajectory; the point motion is large, with an almost uniform coverage of diameters up to 55\% of the image dimensions.
To measure diversity of point motion, we cluster point trajectories within each video ($\approx 30$ trajectories/video) using agglomerative clustering. A large number of clusters would suggest diverse motion of points within the given video.
The distance between two clusters is measured as the minimum distance between any two constituent trajectories, where the trajectory distance is the mean-centered Euclidean distance between non-occluded points. The clusters are recursively merged until all inter-cluster distances are at least 2 pixels, starting with each trajectory assigned to an independent cluster (see Appendix~\ref{a:clustering} for details). We find $\geq$ 85\% videos have $\geq$ 5 clusters, suggesting diverse motion. Finally, a large number of Kinetics-700 action categories (532 unique) are represented in the videos.

\subsection{Evaluation of human annotation quality}

We adopt two approaches to evaluate the human annotation: 1) using simulated groundtruth, and 2) using human inter-rater reliability on real videos. 

\textbf{Simulated Groundtruth}
Kubric~\cite{kubric} contains all the information required to obtain perfect ground truth tracks for any point the annotators might choose.  Therefore, we ask them to annotate 10 Kubric videos with length 2 seconds (50 frames at 25 FPS), where we have known ground truth point tracks.  
With our optical flow track assistance, over 99\% of annotated points are accurate to within 8 pixels, 96\% to within 4 pixels, and 87\% to within 2 pixels.  This is true even though Kubric includes many difficult objects that fall and bounce in unpredictable ways.
In terms of annotation time, optical flow assistance improve annotation speed by 28\%, from 50 minutes per video to 36 minutes per video. This is nontrivial considering the overall video length is only 2 seconds.
See Appendix~\ref{sec:sim_validation} for more details, including a demonstration of the non-trivial improvement from our track assist algorithm.

\textbf{Human Agreement} After a first round of annotation on DAVIS, we perform a second round of annotation using the first frame points from the first round.  That is, we ask different human raters to annotate the same set of points. We then compare the similarity between the human annotations on the same point track using our established metrics. Overall human agrees with 95.5\% on occlusion and 92.5\% on location with a 4 pixel threshold.
On average, there is only a 1.46 pixel difference between two independent human tracks on the same point under 256x256 resolution. See Appendix~\ref{sec:inter_rater_validation} for more details.

\section{Baselines}
Few existing baselines can be applied directly to our task. Therefore, we adapt several state-of-the-art methods which perform different kinds of point tracking with simple extensions.  We expect that ideas from these papers can be better integrated into a more complete point tracking setup, but as baselines, we aim to keep the implementations simple. \textbf{Kubric-VFS-Like}~\cite{kubric}: A baseline for point tracking trained on Kubric using contrastive learning, inspired by VFS~\cite{xu2021rethinking}. 
\textbf{RAFT}~\cite{teed2020raft}: We extend this state-of-the-art optical flow algorithm to multiple frames by integrating the flow from the query point: i.e., we use bilinear interpolation of the flow to update the query point, move to the next frame, and repeat. We handle out-of-bounds points by using the flow from the nearest pixel, and mark points as occluded when they're outside the frame.  \textbf{COTR}~\cite{jiang2021cotr}: This method is intended to be used with larger baselines, so to propagate a query point, we apply this algorithm with the query frame paired with every other frame.  We use cycle consistency inspired by Kubric-VFS-Like~\cite{kubric} to estimate occlusion.  See Appendix~\ref{sec:baselines} for details.

\subsection{TAP-Net}
Overall, we find existing baselines give unsatisfactory performance.  The above baselines don't handle occlusion (RAFT), they deal poorly with deformable objects (COTR), or they exhibit poor transfer performance from synthetic to real data (Kubric-VFS-Like).  This is in part due to the structure of prior point tracking benchmarks: existing optical flow and SFM benchmarks use only frame pairs (thus encouraging algorithms that are too slow for video), and they don't evaluate occlusion estimation.   While numerous prior works contain ideas which likely have a place in solving the TAP task, we here implement a basic baseline  which aims for simplicity while also achieving a fast runtime and reasonable performance. This is the first end-to-end deep learning algorithm we are aware of for tracking any point.

Our approach is inspired by cost volumes~\cite{hosni2012fast,scharstein2002taxonomy,zbontar2016stereo}, which have proven successful for optical flow.  We first compute a dense feature grid for the video, and then compare the features for the query point with the features everywhere else in the video.  Then, given the set of comparisons between a query and another frame, we apply a small neural network which regresses to the point location (trained via Huber loss) and classifies occlusion (trained via cross entropy).

\subsubsection{Cost Volume}

\begin{figure*}[t]
\centering
\includegraphics[width=\linewidth]{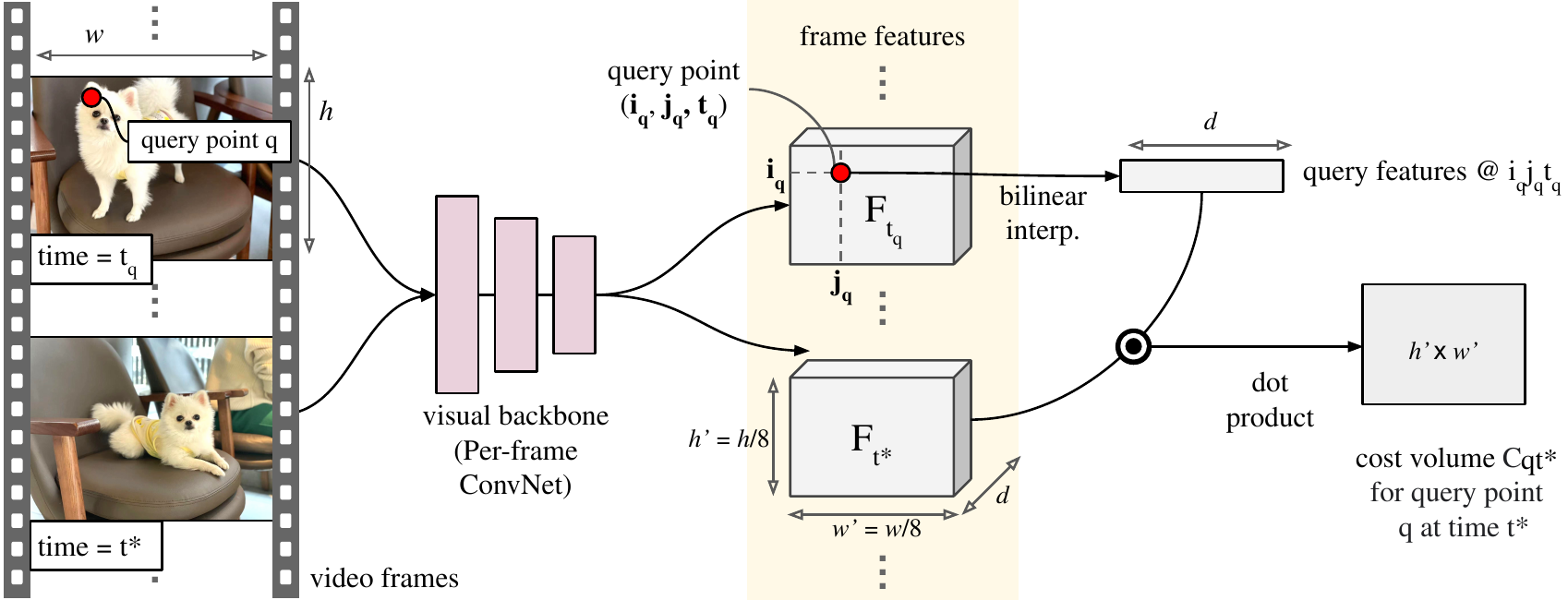}
\caption{\textbf{Cost volume computation for a single query.} Features for all video frames are extracted using a per-frame ConvNet. Features for the given query point location $(i,j)$ in the $t_q$--th frame are then obtained through 2D bilinear interpolation. 
The features are dotted with the spatial features from a different time $t^{*}$ to obtain the corresponding cost volume.}
\label{fig:cost_volume}
\end{figure*}

Any tracking problem involves comparing the query point to other locations in the video in order to find a match. 
Our approach is inspired by cost volumes~\cite{hosni2012fast,scharstein2002taxonomy,zbontar2016stereo}, which have proven successful for optical flow.  We first compute a dense feature grid for the video, and then compare the features for the query point with the features everywhere else in the video.  Then we treat the set of comparisons as if it were a feature itself, and perform more neural network computation on top.  
Cost volumes can detect repeated patterns even when those patterns are unlike those seen in training.  

Figure~\ref{fig:cost_volume} shows the procedure. Given a video, we first compute a feature grid $F$, where $F_{ijt}$ is a $d$-dimensional feature representing the image content at spatial location $i,j$ and time $t$. For this, \changed{a TSM-ResNet-18~\cite{lin2020tsm} with time shifting in only the first two layers, as we find that single-frame feature maps, and feature maps with a larger temporal footprint, tended to perform worse.} 
Given a query point at position $x_q,y_q$ and time $t_q$, we extract a feature to represent it from the feature grid $F_{t}$ via bilinear interpolation on the grid, at position $i_q,j_q,t_q$.  We call the extracted feature $F_{q}$. We then compute the cost volume as a matrix product: i.e., if each feature in the feature map $F$ is shape $d$, then the output cost volume $C_{qijt}=F_{q}^{\top}F_{ijt}$.  Thus, $C_{q}$ is a 3-D tensor, to which we apply a ReLU activation.

\subsubsection{Track Prediction}
\begin{figure*}[t]
\centering
\includegraphics[width=\linewidth]{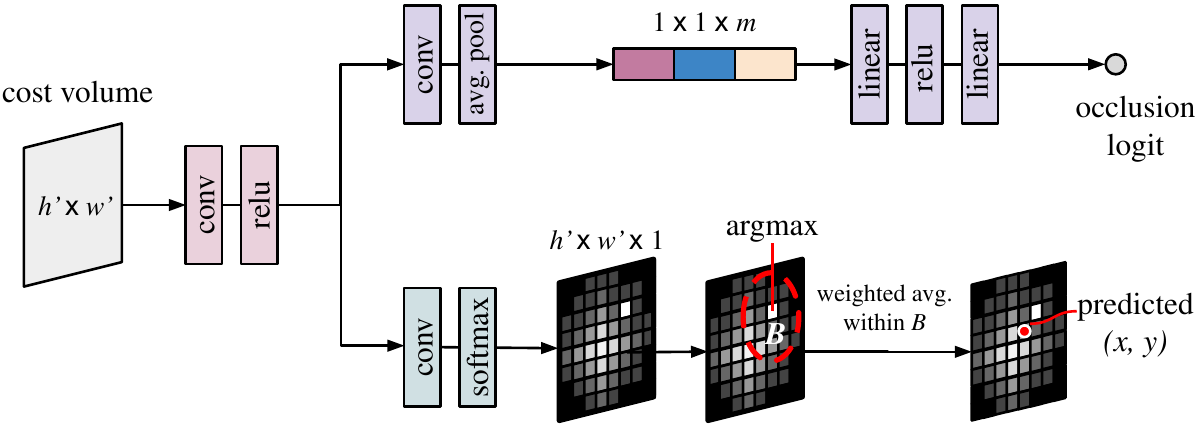}
\caption{\textbf{Inferring position and occlusion from a cost volume.} The cost volume slice (for frame $t^{*}$ and query $q$) is fed into two branches: occlusion and coordinate regression. The occlusion branch collapses the spatial features through pooling before regressing a scalar occlusion logit. The point-regression branch collapses the channels through a Conv layer, applies a spatial softmax, and then applies a soft argmax (weighted average of grid-coordinates within a Euclidean ball $\mathcal{B}$). 
\vspace{-.1em}
}
\label{fig:inference}
\end{figure*}

The next step is to post-process the cost volume associated with the query point, which is done independently for each frame.  
The architecture for one frame is shown in Figure~\ref{fig:inference}.
After a first Conv+ReLU, we split into two branches: one for occlusion inference, and the other for position inference.  \changed{Our occlusion branch uses a Conv (32 units), followed by  spatial average pooling.  This is followed by a linear layer (16 units), a ReLU, and another linear layer which produces a single logit.  }

For position inference, we apply a Conv layer with a single output, followed by a spatial softmax.  This is followed by a soft argmax~\cite{levine2016end}, which means computing the argmax of the heatmap, followed by a spatial average position of the activations within a radius around that argmax location.  

Mathematically, let us assume that $S_{qijt}\in \mathbb{R}$ is the softmax activation at time $t$ and spatial position $i,j$ for query $q$.  Let $G$ be a spatial grid, i.e., $G_{ij}\in \mathbb{R}^{2}$ is the spatial position of $S_{ij}$ in image coordinates.  Finally, let $(\hat{i}_{qt},\hat{j}_{qt})$ be the argmax location of $S_{qt}$.  Then we compute the output position as:
\begin{equation}
    p_{qt}=\frac{\sum_{ij}\mathds{1}\left(||(\hat{i}_{qt},\hat{j}_{qt})-(i,j)||_2<\tau\right)S_{qijt}G_{ij}}{\sum_{ij}\mathds{1}\left(||(\hat{i}_{qt},\hat{j}_{qt})-(i,j)||_2<\tau\right)S_{qijt}} \label{eq:soft_argmax}
\end{equation}
$\tau$ here is a constant, typically equal to 5 grid cells.  Note that some parts of this expression are not differentiable: notably the thresholding and the argmax.  However, the gradients will still point in the right direction, as errors will typically cause the network to shift the overall mass of the softmax toward the ground truth location~\cite{levine2016end}.

\subsubsection{Loss Definition}

Our loss for each query point then has the following form:
\begin{equation}
    L(\hat{p},\hat{o},p^{gt},o^{gt})=\sum_{t}(1-o^{gt}_t)L_H(\hat{p}_t,p^{gt}_t)-\lambda\left[\log(\hat{o})o^{gt}+\log(1-\hat{o})(1-o^{gt})\right]
    \label{eqn:loss}
\end{equation}
Here, $L_H$ is the Huber loss~\cite{huber1992robust}.  $p^{gt}$ is the ground-truth position, and $o^{gt}$ is a (binary) ground truth occlusion.  That is, we treat position as a simple regression problem for frames where the point is visible.  We use a Huber loss because we expect occasional large errors, and we don't want to penalize these excessively.
For occlusions, the loss is a standard cross entropy. $\lambda$ is the trade-off parameter.  For further details, see Appendix~\ref{sec:tapnet_details}.

\section{Results}
\label{sec:experiments}

Table~\ref{tab:sota_comparison} shows the comparison results. TAP-Net outperforms prior baseline methods on all 4 datasets by a large margin, and provides competitive performance with one concurrent work~\cite{harley2022particle}. Kubric-VFS-Like provides somewhat competitive performance, especially on Kubric itself, but its cycle-consistency-based method for detecting occlusions is not particularly effective on real-world data, and the lack of end-to-end training means that its tracks are not very precise. COTR shares similar difficulties on detecting occlusions as it has no built-in method for doing so. Even ignoring occlusion estimation, COTR struggles with moving objects, as it was trained on rigid scenes. COTR sees especially poor performance on the RGB-Stacking dataset; most likely the lack of textures results in degenerate solutions.  RAFT likewise cannot easily detect occlusions, and due to the frame-by-frame nature of tracking, errors tend to accumulate, leading to poor performance.  

Concurrent work PIPs~\cite{harley2022particle}, however, provides perhaps the strongest competition, as it was designed for point tracking in video in the same spirit as our benchmark.  It provides the best performance on \datasetname-DAVIS, although it struggles on \datasetname-Kinetics and \datasetname-RGB-Stacking.  This is likely because PIPs is essentially an online algorithm; given the tracking result for $N$ frames, it performs inference on frames $N$ to $N+8$ assuming that the result on the $N$-th frame is approximately correct, and performs only local searches.  If the track is lost for more than 8 frames (e.g. a long occlusion) or if there is a discontinuity in the sequence (e.g. a cut in the video), then PIPs is likely to fail catastrophically.  DAVIS videos tend to be continuous shots with smooth motion, which means that the `local search' and smoothness inductive biases are helpful; in Kinetics, however, there are more occlusions, cuts, and rapid motion like camera shake, which means that these inductive biases may be more harmful than helpful.  PIPs also struggles on \datasetname-RGB-Stacking, possibly because it also contains long occlusions, but also possibly due to the textureless objects, a problem which might be attributed to the training dataset.  PIPs is trained on FlyingThings3D~\cite{mayer2016large}, and it remains unclear if this is comparable to Kubric for training point tracking.  One advantage of PIPs, however, is that it is straightfoward to higher-resolution videos, although it runs slowly.  We evaluated it on full-resolution DAVIS videos (1080p) and found average jaccard improves from 42.0\% to 48.6\% ($<\delta^{x}_{avg}$ rises from 59.4\% to 66.6\% and occlusion accuracy rises from 82.1\% to 83\%).  While this isn't a fair comparison to other methods, which only have access to lower-resolution videos, we believe that ``full resolution'' evaluation is interesting in its own right, as users will likely want improved accuracy on high-resolution videos.  Therefore, we pose this as a separate challenge for future research.

See the online NeurIPS supplementary material for qualitative examples of tracking results.

\begin{table*}[t]
\resizebox{\linewidth}{!}{ %
\begin{tabular}{l|ccc|ccc|ccc|ccc}
\toprule
 & \multicolumn{3}{c|}{Kinetics} & \multicolumn{3}{c|}{Kubric} & \multicolumn{3}{c|}{DAVIS} & \multicolumn{3}{c}{RGB-Stacking} \\
Method &  AJ & $<\delta^{x}_{avg}$ & OA &  AJ & $<\delta^{x}_{avg}$ & OA &  AJ & $<\delta^{x}_{avg}$ & OA &  AJ & $<\delta^{x}_{avg}$ & OA \\
\midrule
COTR~\cite{jiang2021cotr} & 19.0$^{*}$  & 38.8$^{*}$  & 57.4$^{*}$ &  40.1 & 60.7 & 78.5 & 35.4 & 51.3 & 80.2 & 6.8 & 13.5 & 79.1 \\
Kubric-VFS-Like~\cite{kubric} & 40.5 & 59.0 & 80.0 & 51.9 & 69.8 & 84.6 & 33.1 & 48.5 & 79.4 & 57.9 & 72.6 & \textbf{91.9} \\
RAFT~\cite{teed2020raft} & 34.5 & 52.5 & 79.7 & 41.2 & 58.2 & 86.4 & 30.0 & 46.3 & 79.6 & 44.0 & 58.6 & 90.4 \\
PIPs~\cite{harley2022particle} & 35.1 & 54.8 & 77.1 & 59.1 & 74.8 & 88.6 & \textbf{42.0} & \textbf{59.4} & \textbf{82.1} & 37.3 & 51.0 & 91.6 \\
TAP-Net & \textbf{46.6} & \textbf{60.9} & \textbf{85.0} & \textbf{65.4} & \textbf{77.7} & \textbf{93.0} & 38.4 & 53.1 & 82.3 & \textbf{59.9} & \textbf{72.8} & 90.4 \\
\bottomrule
\end{tabular}
}
\vspace{1em}
\caption{{\bf
Comparison of TAP-Net versus several established methods on \datasetname}.  TAP-Net outperforms all prior works, often by a wide margin. The reported metrics are average Jaccard (AJ), average position accuracy of visible points ($<\delta^{x}_{avg}$), and binary occlusion accuracy (AO) (see section~\ref{sec:evaluation_and_metrics} for details).    For starred entries, the underlying algorithm was not fast enough to run on the full Kinetics dataset in a practical timeframe, so we ran on a random subset.  See Appendix~\ref{sec:baselines} for details.
}
\label{tab:sota_comparison}
\end{table*}

\subsection{Comparison to JHMDB}
Although JHMDB is designed for human pose tracking~\cite{jhuang2013towards}, numerous prior works~\cite{wang2019learning,lai2019self,li2019joint,jabri2020space,xu2021rethinking,vondrick2018tracking,lai2020mast} use it as an evaluation for class-agnostic point tracking. The task is to track 15 human joints given only the joint positions in the first frame.  
This is an ill-posed problem, as the joints are \emph{inside} the object, at some depth which is not known to the algorithm; worse, annotators estimate joint locations even when they are occluded. 
Algorithms typically follow an ad-hoc formulation that converts the points into segments before tracking the segments. 
Despite the dataset's flaws, it remains the best available for point tracking on deformable objects, which explains its popularity (and underscores the need for something better).  
In this section, we aim to demonstrate the generality of \datasetname-Kinetics point tracking by showing that finetuning on it it can improve downstream performance on the much-more-specific (but also highly useful) JHMDB tracking task.
Conversely, we demonstrate the lack of generality of JHMDB by showing that finetuning on JHMDB actually \textit{harms} performance on \datasetname-Kinetics.

We take the full model pretrained on \datasetname-Kubric and fine-tune using the same settings on both Kinetics and JHMDB.  For simplicity, we use no data augmentation on either dataset; both datasets are resized to $256x256$.  Note that the two datasets have similar scale: 928 clips for JHMDB versus $1,280$ for \datasetname-Kinetics.  For each training example, we sample 128 queries from each video uniformly at random from all possible queries (unoccluded tracked points), and feed the corresponding tracks as training examples (note that the same track is often sampled multiple times with different queries).  We train with 1 video per TPU core, and otherwise use the same training setup as for training on Kubric (64 TPU-v3 cores, Adam optimizer).  We train for only 5000 steps (100 warmup steps followed by a cosine schedule) with a much smaller learning rate of $1e-5$ (otherwise the network overfits badly to the relatively small number of points).  For JHMDB, we fine-tune on the full train+val dataset, but we evaluate (after Kinetics fine-tuning) on only the JHMDB val-set in order to be comparable with prior results in the literature.  

Table~\ref{tab:jhmdb_finetuning} shows the performance across all datasets with and without finetuning.  Not that we're particularly interested in transfer (salmon-colored rows); i.e., it's not surprising that performance on Kinetics is best for the model finetuned there, and same for JHMDB, as we train on the evaluation images for both.  However, what's interesting is that our full model obtains $62.3$ PCK@0.1 and $79.8$ PCK@0.2 on JHMDB when trained only on Kubric, but $63.4$ PCK@0.1 and $80.1$ PCK@0.2 after finetuning on Kinetics.  However, the model finetuned on JHMDB obtains $58.6$ $<\delta^{x}_{avg}$ on Kinetics, and $36.4$ Average Jaccard, versus $46.6$ and $60.9$ respectively training purely on synthetic data.  Note the loss of performance on occlusion accuracy and Average Jaccard are perhaps not so surprising, as JHMDB contains no information about occlusion (the evaluation assumes that points which are unoccluded in the first frame are unoccluded in all frames, and we train on all such points).  However, $<\delta^{x}_{avg}$ ignores occlusion, and TAP-Net is still substantially worse with JHMDB finetuning than without it.  We see a consistent downtrend in $<\delta^{x}_{avg}$ (and average Jaccard) throughout training for all datasets, suggesting that this is not merely an issue of overfitting, but rather a fundamental problem with the JHMDB dataset.

\begin{table*}[t]
\resizebox{\linewidth}{!}{ %
\begin{tabular}{l|ccc|ccc|ccc|ccc|cc}
\toprule
 & \multicolumn{3}{c|}{Kinetics} & \multicolumn{3}{c|}{Kubric} & \multicolumn{3}{c|}{DAVIS} & \multicolumn{3}{c|}{RGB-Stacking} & \multicolumn{2}{c}{JHMDB PCK} \\
Method &  AJ & $<\delta^{x}_{avg}$ & OA &  AJ & $<\delta^{x}_{avg}$ & OA &  AJ & $<\delta^{x}_{avg}$ & OA &  AJ & $<\delta^{x}_{avg}$ & OA & @0.1 & @0.2 \\
\midrule
Kubric Only & 46.6 & 60.9 & 85.0 & \textbf{65.4} & \textbf{77.7} & \textbf{93.0} & 38.4 & 53.1 & 82.3 & 59.9 & 72.8 & \textbf{90.4} & 62.3 & 79.8 \\
Fine-tune Kinetics & \textbf{51.6} & \textbf{64.9} & \textbf{90.4} & 59.9 & 73.1 & 90.8 & \textbf{41.5} & \textbf{56.0} & \textbf{82.4} & \textbf{63.2} & \textbf{74.7} & 90.2 & \cellcolor{shade1} 63.4 & \cellcolor{shade1} 80.1 \\
Fine-tune JHMDB & \cellcolor{shade1} 36.4 & \cellcolor{shade1} 58.6 & \cellcolor{shade1} 71.3 & 51.4 & 69.9 & 83.9 & 31.6 & 49.0 & 77.0 & 53.0 & 69.1 & 90.9 & \textbf{71.3} & \textbf{87.7} \\
\bottomrule
\end{tabular}
}
\vspace{1em}
\caption{{\bf
TAP-Net performance after fine-tuning} for all tasks, compared to no fine-tuning.  The relevant cross-dataset transfer results are highlighted.  We see that fine-tuning on JHMDB harms performance on all datasets except JHMDB by a large margin, but finetuning on Kinetics improves performance on JHMDB, demonstrating the domain transfer properties of the TAP task when the dataset contains arbitrary points tracked on a diverse YouTube dataset. The cross-dataset performance of particular interest is highlighted.
}
\label{tab:jhmdb_finetuning}
\end{table*}

\section{Conclusions}
In this paper we introduce the problem of \textit{Tracking Any Point} (TAP) in a given video, as well as the \datasetname~dataset to spur progress in this under-studied domain.
By training on synthetic data, a straightforward network TAP-Net performs better on our benchmark than prior methods.
TAP still has limitations: for instance, we cannot handle liquids or transparent objects, and for real data, \changed{annotators cannot be perfect, as they are limited to textured points and even then may make occasional errors due to carelessness}.
We believe that the ethical concerns of our dataset are minimal; \changed{however, our real data comes from existing public sources, meaning PII and biases must be treated with care to ensure no harms to anyone in those sources, as well as fairness of the final algorithm}.
For our largest dataset, Kinetics, we will follow its approach of removing videos that are taken down from YouTube to ensure consent and issues with PII. 
Kinetics has also been vetted for biases~\cite{carreira2017quo}; training synthetic data should further minimize biases that might be learned by the algorithm. 
Advancements in TAP will potentially bring solutions to many interesting challenges, e.g., better handling of dynamic or deformable objects in SFM \cite{schoenberger2016sfm}, and allowing the semantic keypoint-based methods~\cite{florence2018dense,manuelli2019kpam,vecerik2020s3k,manuelli2020keypoints} often employed in robotic object manipulation to be applied to generic objects, independent of their class.  

\section{Acknowledgements}
We wish to thank Klaus Greff and Andrea Tagliasacchi for discussions on Kubric; Alina Kuznetsova, Yiwen Luo, Aakrati Talati and Lily Pagan for helping us with the human annotation tool; Muqthar Mohammad, Mahesh Maddinala, Vijay Vibha Tumala, Yilin Gao, and Shivamohan Garlapati for help with organizing the human annotation effort;  Adam Harley for help with PIPs; Jon Scholz, Mel Vecerik, Junlin Zhang, Karel Lenc, Viorica Patraucean, Alex Frechette and Dima Damen for helpful discussions, advice, and side projects that shaped this paper, and finally Nando de Freitas for continuously supporting the research effort.

\clearpage

\bibliographystyle{splncs04}
\bibliography{egbib}

\appendix
\begin{figure*}[hp]
\centering
\includegraphics[width=.99\linewidth]{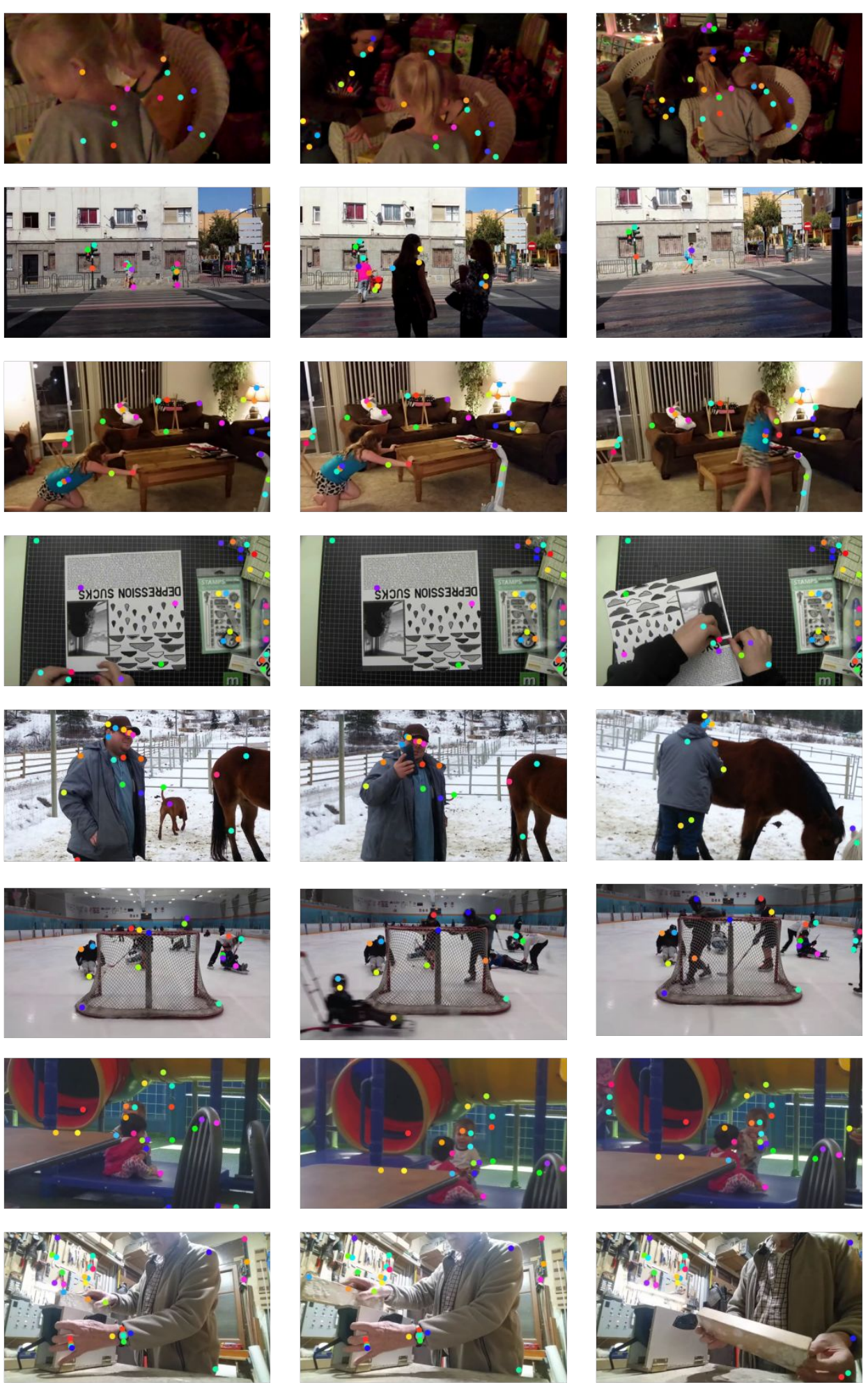}
\caption{{\bf Examples from TAP-Vid-Kinetics}: In this figure we show a few examples of TAP-Vid-Kinetics annotations with 3 frames per example.
}
\label{fig:tap-vid-kinetics}
\end{figure*}

\begin{figure*}[hp]
\centering
\includegraphics[width=.99\linewidth]{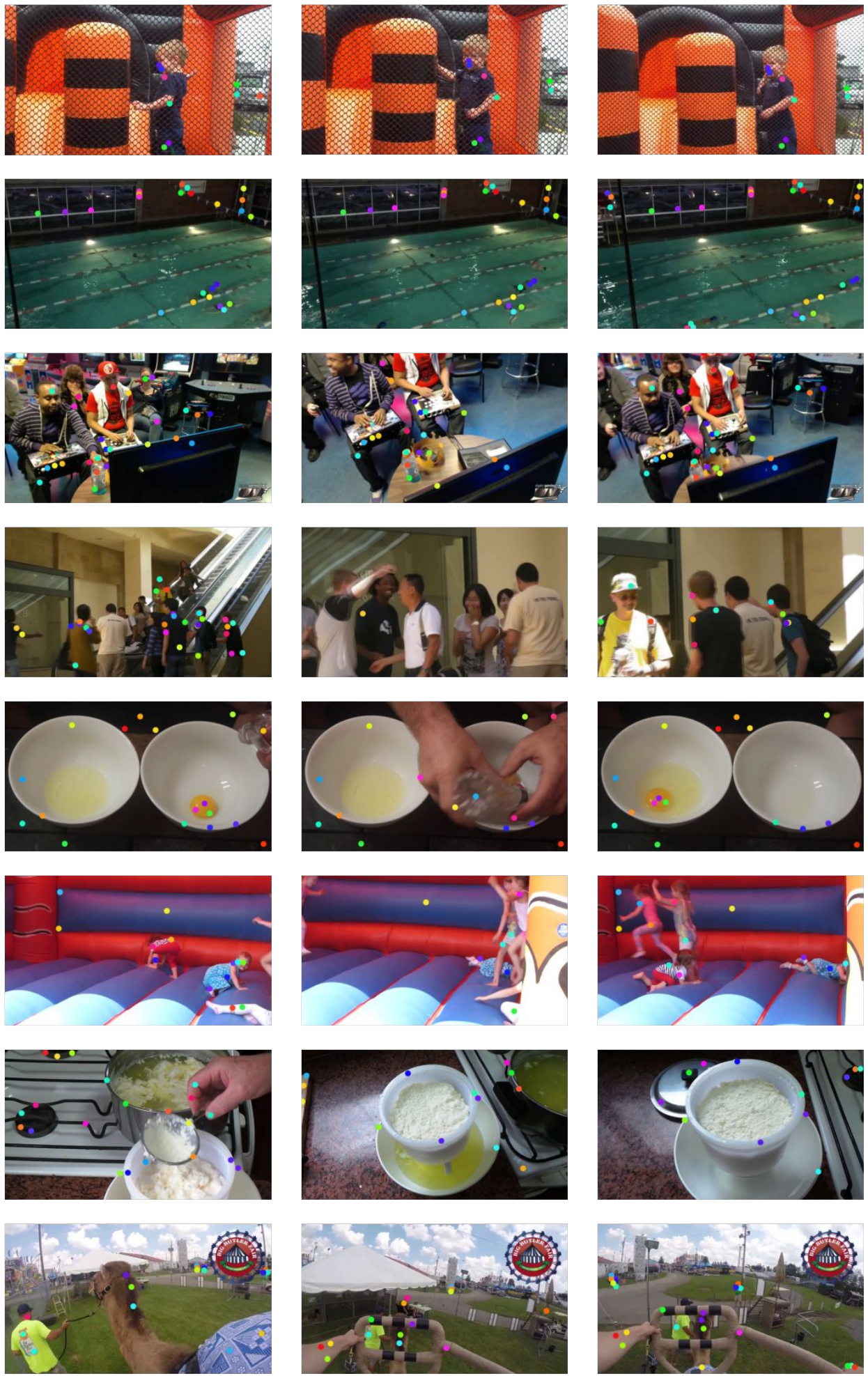}
\caption{{\bf Examples from TAP-Vid-Kinetics}: In this figure we show a few examples of TAP-Vid-Kinetics annotations with 3 frames per example. 
}
\label{fig:tap-vid-kinetics2}
\end{figure*}

\begin{figure*}[hp]
\centering
\includegraphics[width=.99\linewidth]{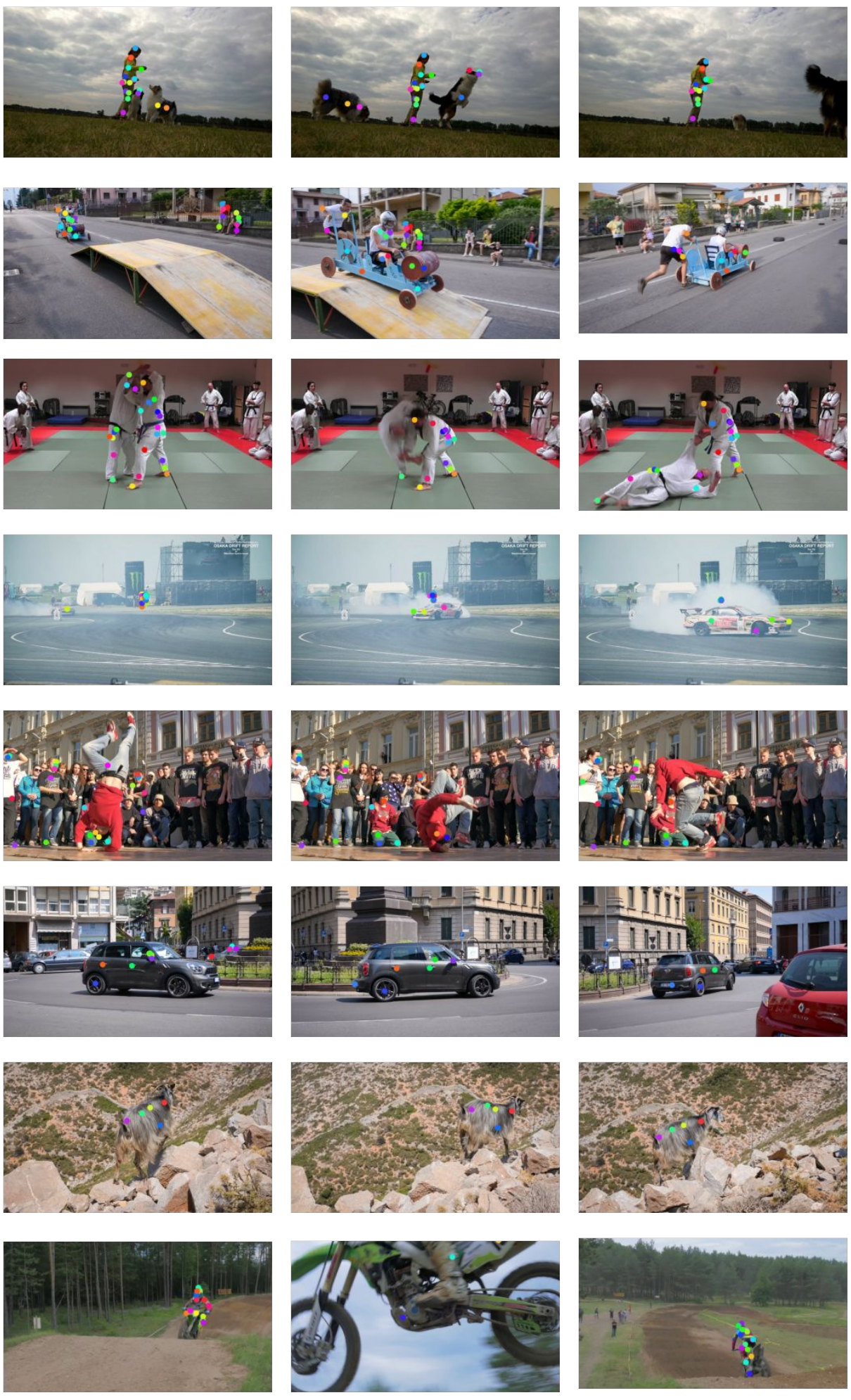}
\caption{{\bf Examples from TAP-Vid-Davis}: In this figure we show examples of TAP-Vid-Davis annotations with 3 frames per example.
}
\label{fig:tap-vid-davis}
\end{figure*}

\begin{figure*}[hp]
\vspace{3cm}
\centering
\includegraphics[width=.99\linewidth]{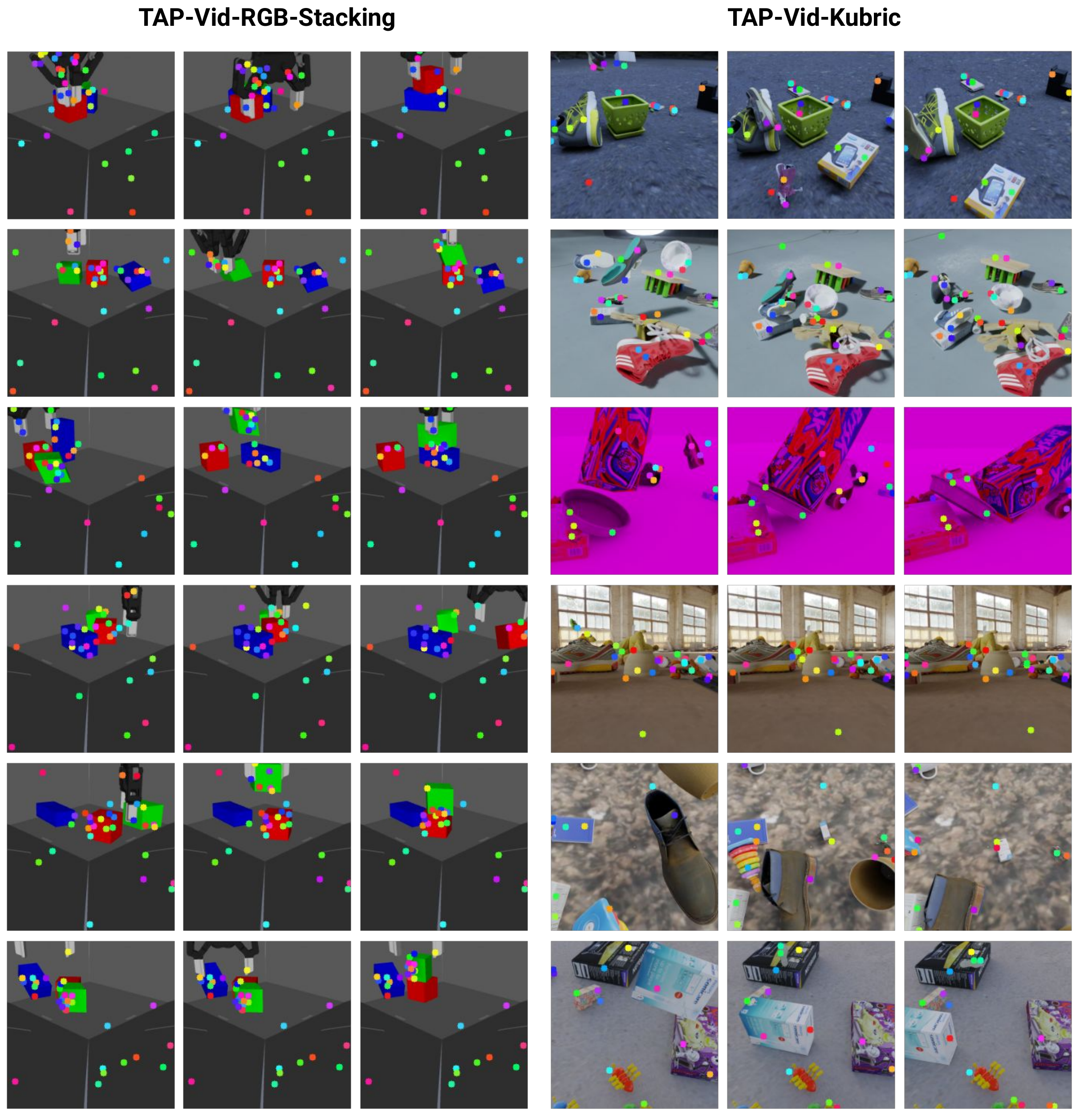}
\caption{{\bf Examples from TAP-Vid-Kubric and TAP-Vid-RGB-Stacking}: In this figure we show examples of TAP-Vid-Kubric and TAP-Vid-RGB-Stacking annotations with 3 frames per example.
}\label{fig:tap-vid-robotics}
\end{figure*}

\section{Dataset Examples}\label{s:dset-ex}
To provide more context and information to the reader, we expand Figure 2 in the paper providing many examples per dataset. We provide examples of TAP-Vid-Kinetics in Figures~\ref{fig:tap-vid-kinetics} and ~\ref{fig:tap-vid-kinetics2}, TAP-Vid-Davis in Figure~\ref{fig:tap-vid-davis} and TAP-Vid-Kubric and TAP-Vid-RGB-Stacking in Figure~\ref{fig:tap-vid-robotics}. These figures illustrate the diversity of the tracked points across all the different datasets. Figures ~\ref{fig:tap-vid-kinetics}, ~\ref{fig:tap-vid-kinetics2} and \ref{fig:tap-vid-davis} show how we are tracking points beyond the most salient objects in the video; the vast majority of tracked points do not correspond to the semantic keypoints of existing datasets. 
The website contains even more examples in video format, under the `Visualized Ground Truth Examples' heading.

\section{Supplementary Videos, Code, and Website}

The website for this project \url{https://github.com/deepmind/tapnet} contains the following:
\begin{enumerate}
    \item \textbf{Ground Truth Point Tracking Annotations.} Video examples of groundtruth point annotations in our datasets, as summarized in Section~\ref{s:dset-ex} above.
    \item \textbf{Dataset Download Links and Processing.} Download links for the \datasetname-\{Kinetics, DAVIS, RGB-Stacking\} datasets, and instructions for processing and aligning raw Kinetics videos to the annotations and for running the visualization scripts. It also includes licensing information.
     \item \textbf{Improving Human Point Annotation with Flow-Based Tracker.} Comparison of human point annotation quality on the DAVIS dataset with and without optical flow track assistance.
    
\end{enumerate}

\section{Annotation Instructions and Workflow}
\label{sec:annotation_instructions}
\begin{figure}[t]
\centering
\MakeFramed{\advance\hsize-\width\FrameRestore}%
\noindent\hspace{-4.55pt}%
\begin{adjustwidth}{}{7pt}%
\vspace{2pt}\vspace{2pt}%
\begin{enumerate}
\item View the video once. Choose a point that appears on the object (specified by the given bounding-box) for the very first time. Prefer locations such that the point marker plotted at the chosen location should have its center and periphery both completely inside the object region. Try to spread out the points on different parts and label them with part tags if possible (i.e., human hand, human head, car wheel etc.).

\item Find the last frame after which the point gets occluded (e.g., because the object moves outside of the frame) and end the track. Then if it becomes visible again, and start a new track to follow it under the same tag, repeating this process until the end of the video.

\item Wait for the optical flow assist algorithmic annotations to adjust based on the manual input. Once the algorithmic annotations are available, add manual annotations until both manual and automated annotations are consistent with the schematic below:

\begin{minipage}{\linewidth}
\centering
\includegraphics[width=.8\linewidth]{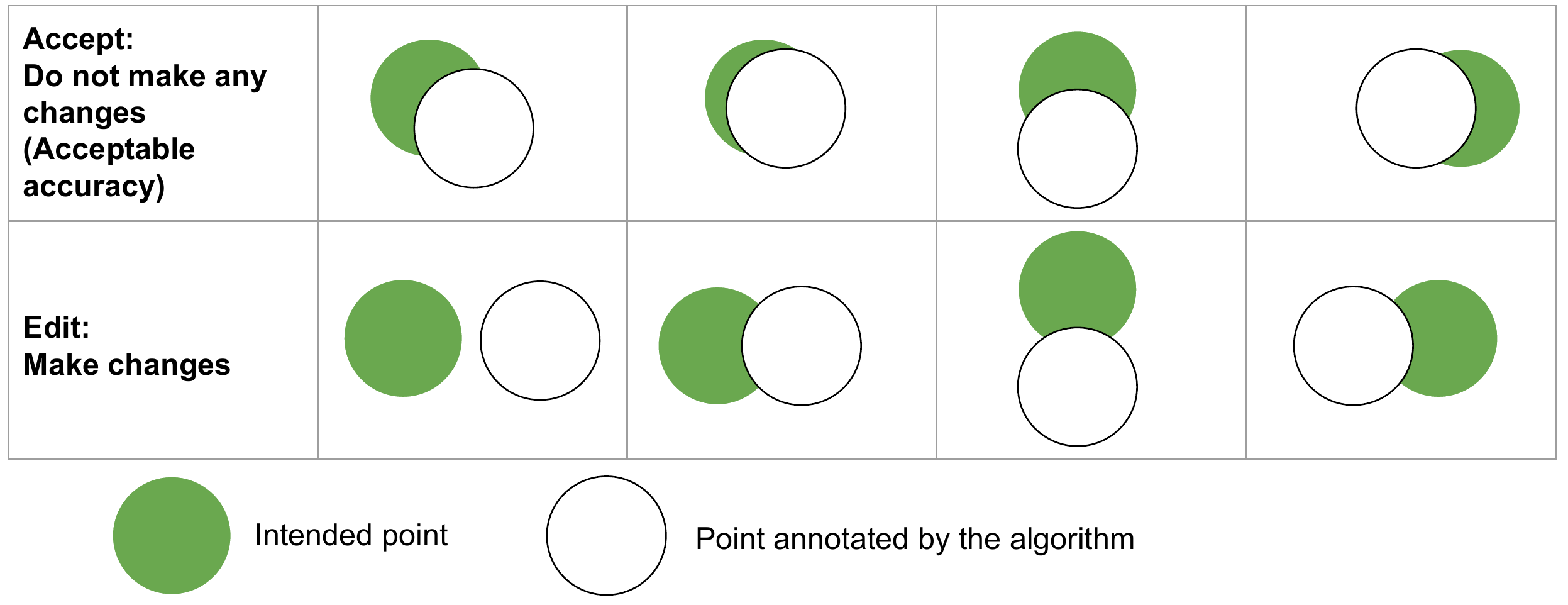}
\end{minipage}

\item Go through the video again to identify any interpolated points in the un-annotated frames with obvious position errors. Correct the error by manually adding a point.
Wait a few seconds for the assist algorithm to update the point tracks.
Continue the process of reducing errors in the remaining un-annotated frames.

\item Once satisfied with the point track in all frames where it is visible, choose another point and repeat the process from the beginning.
\end{enumerate}
\vspace{2pt}\end{adjustwidth}\endMakeFramed%
\caption{{\bf Representative annotator guidelines.} (Minor variations were used as we saw results and gave feedback). }
\label{fig:instructions}
\end{figure}

Figure~\ref{fig:instructions} summarizes the guidelines given to the annotators to encourage acquisition of high-quality point tracks.
Note, the guidelines do not limit the scope of the kind or type of points the annotators can select (e.g., a specific object category etc.); hence, \emph{any} point which the annotators can track reliably is admissible, resulting in a diverse set of point trajectories in the dataset.
The annotators are encouraged to leverage the optical flow track assist algorithm to help improve their accuracy and productivity, \changed{although they have the option to turn it off if it's not helpful, resulting in a fallback to linear interpolation}. A visual representation of acceptable discrepancy between manual and automated point selections is provided to help the annotators decide when to intervene by adding another point.

\section{Simulated Dataset Generation}
\label{sec:sim_data}

For synthetic data, obtaining point tracks is conceptually straightforward, as the simulator typically knows the shape and position of all objects.  However, most modern simulators don't directly expose point tracks, so we must compute them.  Since both Kubric and RGB-Stacking contain only rigid objects, we therefore obtain point tracks in the following manner.  We first choose query points in pixel space, which ensures that the query is visible. Then we find the location of that point in the object's local coordinate frame, using either coordinate maps (available for Kubric objects) or by back-projecting using the depth map and applying the inverse object transformation.  Next, we compute the 3D location in world coordinates for every other frame using the known transformation of each object.  Finally we project the point to the image plane.  To compute occlusion, we compare the computed depth at the target frame with the depth map rendered from the simulator.  Specifically, if the point is behind the depth map (the max depth of the 4 nearest pixels to the reprojected point) by a sufficient margin (1\%), we mark it as occluded.  

For Kubric, we sample exactly 256 query points from each video, and we try to have the same number of points for every object, including the background. However, this can lead to far too many points for very small objects, so we limit sampling to 0.16\% of the total pixels on each object.  There may then not be enough pixels on the smallest objects to sample the same number for each object.  Therefore, we sort objects by the number of pixels, and sample iteratively starting with the smallest: if there are $K$ objects, and the smallest object has only $P$ pixels, then we sample $N$ pixels from this smallest object, where $N=\lfloor\mbox{min}(P*.0016 , 256/K)\rfloor$.  Then we reduce $256$ by $N$, reduce $K$ by 1, and repeat until we've sampled points from all objects.  This ensures the most balanced set of points possible without having too many points per object.

For \datasetname-RGB-Stacking, we sample exactly 600 query points uniformly from the first frame for each video and tracked them via the described method above throughout the video. Then we sample 30 points per video such that 20 of them are from moving objects (i.e.\ tracks with significant motion) and 10 points are from static objects/background (i.e.\ tracks with minimal or no motion).

\section{Dataset Statistics: Agglomerative Clustering of Trajectories}\label{a:clustering}
To assess diversity of point motion in our annotations, we cluster the point trajectories in each labelled video (approx.\ 30 trajectories per video). Agglomerative clustering is performed starting with assigning each trajectory to an independent cluster and recursively merging the clusters until all the inter-cluster distance are greater than a threshold (set to 2 pixels). The cluster distance between two given clusters is the minimum distance between any two trajectories in the two clusters. The distance between trajectories is computed as the mean-centered Euclidean distance between non-occluded trajectory points; this distance is assumed to be undefined (i.e., not counted when comparing clusters) if there are less than 10 frames of overlap. Listing~\ref{code:cluster} gives a Python-pseudocode implementation of the above.

\begin{listing}
\begin{minted}[bgcolor=formalshade]{python}
import numpy as np

def track_dist(t1, t2):
  vis = np.logical_and(t1.raw_vis, t2.raw_vis)
  if np.sum(vis.astype(np.float32)) <= 10:
    return float('inf')
  xy1 = t1.t_xy[vis]
  xy2 = t2.t_xy[vis]
  offset = np.mean(xy1 - xy2, axis=0, keepdims=True)
  xy2 = xy2 + offset
  dist = np.sqrt(np.sum(np.square(xy1 - xy2), axis=-1))
  return np.mean(dist)

def cluster_dist(c1, c2):
  min_dist = float('inf')
  for t1 in c1.children:
    for t2 in c2.children:
      d = track_dist(t1, t2)
      min_dist = min(d, min_dist)
  return min_dist

def all_dists(clusters):
  n = len(clusters)
  dists = np.zeros((n, n)) + float('inf')
  for i in range(n-1):
    c_i = clusters[i]
    for j in range(i+1, n):
      c_j = clusters[j]
      if c_i and c_j are not merged:
        dists[i, j] = cluster_dist(clusters[i], clusters[j])
  return dists

def greedy_cluster(xy, occ, dist_thresh=2/256):
  """
  Agglomerative trajectory clustering.

  xy: [N, T, 2] shaped batch of N trajectory coordinates.
  occ: [N, T] shaped batch of N trajectory occlusions.
  dist_thresh: float, distance threshold for merging clusters.
  """
  # initialize with one cluster per trajectory
  clusters = []
  for i in range(xy.shape[0]):
    clusters.append((xy[i], occ[i]))
  # compute all NxN pairwise trajectory distances
  dists = all_dists(clusters)
  while np.min(dists) < dist_thresh:
    min_ij = np.argmin(dists)
    i, j = np.unravel_index(min_ij, dists.shape)
    # merge trajectories i and j (impl omitted)
    ...
    # re-compute trajectory distances:
    dists = all_dists(clusters)
  return clusters  
\end{minted}
\caption{{\bf Agglomerative clustering of trajectories.}}
\label{code:cluster}
\end{listing}

\section{Quantitative Validation of the Human Annotation Procedure}

\subsection{Validation on Simulated Videos}
\label{sec:sim_validation}

While it is difficult to validate our annotation procedure on real data, we can easily do so on synthetic data where the ground truth is known.  We annotate videos with length 2-seconds (24-frame) from the Kubric dataset~\cite{kubric}, where we have known ground truth point tracks.  Annotators are instructed to label 3 points on any object up to a maximum of 10 objects (up to 30 points per video).  We instruct annotators to choose points which are visible in the first frame, and treat those as queries for computing ground truth tracks.

Because it is a graphics package, Kubric contains all the information required to obtain perfect ground-truth tracks for any point the annotators choose.  We can compare the annotated tracks to the ground truth using the same metrics proposed in the original point tracking task.

To make the videos more representative of realistic annotation challenges, we simulate camera jitter (not present in MOVi-E), which is a common phenomenon in many real-world videos.  We also increase the framerate from 12 FPS to 25, to match our real videos.  All generated videos are included in our supplementary html files.  Annotators are instructed to label 3 points on any object up to a maximum of 10 objects (up to 30 points per video).  We can compare the annotated tracks to the ground truth.  In both cases, annotation was performed at $1024\times 1024$ resolution, although the evaluation is still done at $256\times 256$ pixels to be consistent with our evaluation procedure.  The most intuitive method is the fraction of points with error less than a given threshold, although for completeness we include all metrics proposed for evaluation.  

Table~\ref{tab:annotation_acc} gives our full results, demonstrating the accuracy of our annotators given the optical flow track assist algorithm.
Table~\ref{tab:annotation_acc} also compares results with and without the proposed optical flow track assist algorithm.  We see non-trivial improvements in accuracy with the proposed tool, especially on the metrics emphasizing higher levels of precision: for instance, the error rate for pixels  $< \delta^{1}$ (2 pixels away) has been cut by almost half, and for $< \delta^{0}$ (1 pixel away) it has increased by over 25\% in absolute terms.  This is true even though annotation was done much faster: time was reduced from 50 minutes per video to 36 minutes per video by using the OF track assist.

\begin{table*}[t]
\resizebox{\linewidth}{!}{ %
\begin{tabular}{l|ccc|ccccc|ccccc}
\toprule
method &  AJ & $<\delta^{x}_{avg}$ & OA & Jac. $\delta^{0}$ & Jac. $\delta^{1}$ & Jac. $\delta^{2}$ & Jac. $\delta^{3}$ & Jac. $\delta^{4}$ & $<\delta^{0}$ & $<\delta^{1}$ & $<\delta^{2}$ & $<\delta^{3}$ & $<\delta^{4}$\\
\midrule
No OF  & 74.7 & 83.1 & 97.4 & 25.8 & 64.2 & 90.5 & 96.2 & 96.9 & 41.2 & 78.8 & 96.3 & 99.5 & 100.0 \\
With OF & 81.4 & 90.0 & 96.9 & 49.5 & 76.1 & 90.0 & 95.2 & 96.0 & 66.6 & 87.5 & 96.2 & 99.5 & 100.0 \\

\bottomrule
\end{tabular}
}
\vspace{1em}
\caption{
Annotation accuracy with and without the optical flow track assist algorithm.  
Jac.~$\delta^{x}$ is the Jaccard metric measuring both occlusion estimation and point accuracy, with a threshold of $\delta^{x}$;
AJ is the Average Jaccard across $x$ between $0$ and $4$.  
$<\delta^{x}$ is the fraction of points not occluded in the ground truth for which the prediction is less than $\delta^{x}$, and~$<\delta^{x}_{avg}$ is the average across $x$ between $0$ and $4$.
Occlusion Accuracy is denoted with OA. 
We set $\delta=2$.
}
\label{tab:annotation_acc}
\end{table*}

\subsection{Validation on Inter-Rater Agreement}
\label{sec:inter_rater_validation}

We further conduct inter-human agreement studies on the DAVIS point track videos. For the total of 650 points in 30 videos, we select the first-frame points from the first round of annotation, and ask 2 human raters annotate the following frames for each point. We then compare the similarity between the 2 human annotations using our established metrics. Table~\ref{tab:human_agreement} shows the human agreement results. On average, human on occlusion with 95.5\% accuracy, and 92.5\% on location with a 4 pixel threshold. Note that the metrics are evaluated under 256x256 resolution.

\begin{table*}[t]
\begin{center}
\vspace{.5em}
\resizebox{0.5\linewidth}{!}{ %

\begin{tabular}{c|ccccc}
\toprule
OA & $<\delta^{0}$ & $<\delta^{1}$ & $<\delta^{2}$ & $<\delta^{3}$ & $<\delta^{4}$\\
\midrule
95.5 & 51.8 & 78.3 & 92.5 & 98.7 & 100.0 \\

\bottomrule
\end{tabular}
}
\end{center}
\caption{
Inter-rater agreement on the same set of point tracks for DAVIS.  
$<\delta^{x}$ is the fraction of points not occluded in the ground truth for which the prediction is less than $\delta^{x}$, $x$ between $0$ and $4$.
We set $\delta=2$.
Occlusion Accuracy is denoted with OA.
}
\label{tab:human_agreement}
\end{table*}

\section{TAP-Net Implementation Details}
\label{sec:tapnet_details}

After computing the cost volume, we feed the output to a Conv layer with stride 1, a $3\times3$ receptive field, and 16 hidden units, followed by a ReLU.  

\changed{The occlusion branch takes this output and applies a convolution with stride 2, $3\times3$ receptive field, and 32 hidden units.  This is followed by average pooling, a linear layer with 16 hidden units, ReLU, and a final linear layer for output.}

The prediction branch applies a Conv layer with stride 1, $3\times3$ receptive field, and 1 hidden unit, which produces the input for the softmax, which is then used in the soft argmax above.

When computing the Huber loss, we first rescale both the prediction and ground truth coordinates to be in the range [-1,1].  For each coordinate, we compute the Euclidean distance $||(\hat{x}_{qt},\hat{y}_{qt}) - (x_{qt}^{gt},y_{qt}^{gt})||_2=\sqrt{(x_{qt}-\hat{x}^{gt}_{qt})^{2}+(y_{qt}-\hat{y}^{gt}_{qt})^{2})}$.  Then we apply a Huber loss with a threshold $1/32$ (which corresponds to 4 pixels error in the $256\times256$ images we train on).  The Huber loss is scaled much smaller than the sigmoid cross entropy, so we set $\lambda=100$. 

\changed{Our TSM-ResNet-18 follows ResNet-18: i.e., it has no `bottleneck' layers, so each residual unit is 2 convolutional layers with a $3\times 3$ spatial receptive field.  Each `layer' contains 2 units. It uses no temporal downsampling, meaning that each output has a temporal receptive field of 9 frames.  The final output has 256 channels. } 

\subsection{Training}
We use a cosine learning rate schedule with a peak rate of $2e^{-3}$ and 5,000 warmup steps.  
We train on 64 TPU-v3 cores with a batch size of 4 MOVi-E 24-frame videos per core.  Each model trains for 50,000 steps.
We use an Adam optimizer with $\beta_1=.9$ and $\beta_2=.95$.  We also use weight decay of $1e^{-2}$, applied after the Adam rescaling (i.e.\ Adam-W).  
We use cross-replica batch norm in the ResNet/TSM-ResNet backbone, but don't use batch norm anywhere in or after the cost volume computation.

\changed{We performed minimal hyperparameter tuning by observing transfer performance primarily on DAVIS, tuning learning rate, model size, and optimizer parameters in separate sweeps.  Each experiment requires roughly 24 hours.  Roughly 25 full-scale experiments were required to arrive at the final model.}

\subsection{Model Ablations}
We next ablate our design decisions.  For this, we use \datasetname-Kinetics, the largest and most realistic dataset available to us.  We follow the same training setup as TAP-Net for all experiments.

\begin{table*}[t]
\resizebox{\linewidth}{!}{ %
\begin{tabular}{l|ccc|ccccc|ccccc}
\toprule
Method &  AJ & $<\delta^{x}_{avg}$ & OA & Jac. $\delta^{0}$ & Jac. $\delta^{1}$ & Jac. $\delta^{2}$ & Jac. $\delta^{3}$ & Jac. $\delta^{4}$ & $<\delta^{0}$ & $<\delta^{1}$ & $<\delta^{2}$ & $<\delta^{3}$ & $<\delta^{4}$\\
\midrule
Full Model & 46.6 & 60.9 & 85.0 & 11.3 & 31.6 & 53.3 & 65.8 & 71.2 & 19.8 & 46.4 & 69.3 & 81.7 & 87.4 \\
ResNet-18 & 46.0 & 60.7 & 85.8 & 9.6 & 29.5 & 52.7 & 66.2 & 71.9 & 17.6 & 45.0 & 69.8 & 82.8 & 88.6 \\
Full TSM & 45.4 & 60.1 & 83.7 & 10.8 & 31.0 & 52.1 & 64.0 & 69.1 & 19.2 & 45.9 & 68.5 & 80.6 & 86.4 \\
No Soft Argmax & 45.8 & 59.0 & 84.6 & 10.4 & 30.8 & 52.6 & 64.8 & 70.2 & 18.2 & 44.8 & 67.4 & 79.1 & 85.3 \\
\bottomrule
\end{tabular}
}
\vspace{1em}
\caption{
\changed{Ablation of the architectural choices for our model on \datasetname-Kinetics.  ResNet-18 has no temporal receptive field in the backbone, whereas Full TSM uses time-shifting at every layer, giving a large temporal receptive field.  No Soft Argmax replaces the soft argmax with a simple MLP.  }
}
\label{tab:architecture}
\end{table*}

\changed{We first ablate our backbone network.  Recall that we use a TSM-ResNet-18, with shifting only in the first two layers.  Thus we try TSM-ResNet-18, which uses time shifting at every layer, and ResNet-18, without time shifting.  We also tried replacing the soft argmax operation with a 2-layer MLP regressor (128 hidden units, ReLU activations) that operates directly on the output heatmap.  }

\changed{We see that all of these changes harm the performance to some extent.  Surprisingly, TSM-ResNet performs worse despite allowing the network to integrate across more frames; one possible interpretation is that it gives the network an opportunity to memorize motions in the Kubric dataset.  On the other hand, giving no temporal information is also somewhat worse, particularly for the Average Jaccard metric (though not for occlusion estimation), suggesting that this method makes errors on occlusion estimation even when the location is correct and vice-versa.  One possible interpretation is that with a small amount of temporal information, the network can do a better job of segmenting the object of interest, which allows the network to track and estimate occlusion for the body as a whole.  Allowing this kind of reasoning while preventing memorization of motion patterns presents an interesting challenge for future research.  
The soft argmax is also important, possibly because it enforces that the network performs matching rather than memorizing motion patterns.  
}

\section{Query sampling strategies}

\changed{Recall that our prediction algorithms take as input \textit{query points}, which specify which point needs to be tracked.  For most of this work, we assume that all points are treated equally: any point along a track may be sampled as a query.  In practice, we sample queries in a \textit{strided} fashion: every 5 frames starting at frame 0, we sample a query for every visible point.  This means we may have multiple queries with the same target output trajectory.  An alternative approach which is potentially relevant in an online setting is to start with the \textit{first} frame where the point is visible, and track only into the future (during evaluation, we ignore the predictions for frames earlier than the query frame, as the algorithm can easily assume that these points are occluded, and we don't want to encourage researchers to hardcode this).  In practice, this setting is harder, because on average, the query frame will be farther from each output frame.  However, we include both in order to facilitate comparisons with online methods.  Table~\ref{tab:query_method} shows the comparison for the three relevant datasets.  Kubric is not included, as it uses a different, randomized sampling strategy detailed in the original paper~\cite{kubric}.  Note that query sampling is a potential area of improvement for this dataset: some queries may be ambiguous if they are, e.g., behind translucent objects or very close to occlusion boundaries.  However, we leave the problem of query sampling to future work.}

\begin{table*}[t]
\begin{center}

\resizebox{.8\linewidth}{!}{ %
\begin{tabular}{l|ccc|ccc|ccc}
\toprule
Query method & \multicolumn{3}{c|}{Kinetics} & \multicolumn{3}{c|}{DAVIS} & \multicolumn{3}{c}{RGB-Stacking} \\
Strided & 46.6 & 60.9 & 85.0 & 38.4 & 53.1 & 82.3 & 59.9 & 72.8 & 90.4 \\
First & 38.5 & 54.4 & 80.6 & 33.0 & 48.6 & 78.8 & 53.5 & 68.1 & 86.3 \\
\bottomrule
\end{tabular}
}
\end{center}
\vspace{-1em}
\caption{
Comparison of performance for the same output tracks, but with different query points.  `First' refers to using only the first visible point in each trajectory as a query, whereas `Strided' means sampling multiple query points per trajectory at a stride of 5 frames.
}
\label{tab:query_method}
\end{table*}

\section{Baselines}
\label{sec:baselines}
\paragraph{Kubric-VFS-Like} We first evaluate a point tracking algorithm inspired by VFS~\cite{xu2021rethinking} released as part of Kubric~\cite{kubric}, and the only algorithm which can be applied directly to our setup.  This algorithm uses a contrastive loss to learn features, contrasting points along the trajectory with points off of the trajectory.  At test time, this algorithm computes dot products between the query features and the features at every other frame, before applying a spatial soft argmax similar to the one described in equation~\ref{eq:soft_argmax} for each frame.  It finally estimates occlusion via cycle consistency: i.e., by computing correspondence backward from the target frame to the query frame, and seeing if the correspondence is the same.  \changed{Note that our metrics don't exactly match those from the original paper~\cite{kubric}; the codebase is the same, but we used a ResNet-18 backbone to match this paper, and unlike the original work, we include the query points in the evaluation.}  

\paragraph{RAFT~\cite{teed2020raft}} This algorithm performs near state-of-the-art for optical flow estimation, i.e., tracking points between individual frames.  We extend this to multiple frames by integrating the flow from the query point: i.e., we use bilinear interpolation of the flow estimate to update the query point, and then move to the next frame and repeat.  If the point is outside the frame, we use the flow value at the nearest visible pixel.  While RAFT is extremely accurate over short time scales, this approach has some obvious problems with long time scales. First, it provides no simple way to deal with occlusions (we simply mark points as occluded if they're outside the frame). Furthermore, it has no way to recover from errors, so slight inaccuracies on each frame tend to accumulate over time even when the point remains visible.

\paragraph{COTR~\cite{jiang2021cotr}} Like RAFT, COTR operates on pairs of images; however, it was designed to handle substantially more motion between images.  The underlying architecture uses a transformer to find global scene alignment, followed by more local refinement steps to ensure high accuracy.  It is trained on MegaDepth~\cite{li2018megadepth}, which contains real-world scenes and uses reconstruction to get ground-truth correspondence.  Due to this robustness, we use a different strategy from RAFT to apply COTR to videos: we compare the query frame to every other frame in the video, and find correspondences directly.  COTR has no simple mechanism for detecting occlusions, so  we apply the cycle-consistency strategy proposed in Kubric-VFS-Like: given correspondences between the query and target frame, we take the points in the target frame and find their correspondence in the query image.  If the distance to the original query is greater than 48 pixels, we mark the point as occluded.  Note that it's very expensive to run the model in this way: if there's queries on every frame, then we need to run COTR on every frame pair.  Each pair takes 1 second with 30 points.  Thus, for evaluation, we simplify our benchmark and take only queries from a single frame; thus these numbers are not precise, but good enough to get an idea of COTR's performance on our benchmarks.

\paragraph{PIPs~\cite{harley2022particle}} We ran Persistent Independent Particles (PIPs) following the "chaining" workflow, as the bulk of the algorithm is intended to operate on 8-frame segments.  PIPs extracts features for both the query point and the rest of the video using a ConvNet, similar to TAP-Net.  With the chaining workflow, tracks on each segment are estimated one-by-one, with the final point of each segment used to initialize the subsequent segment.  Given an initial position estimate for a segment, it refines both the query features and the output trajectory using an MLP-Mixer applied iteratively.  Similar to TAP-Net, the input for this refinement network includes the dot product between the query features and the features in the other frames, although in the case of PIPs, the dot products come from a local neighborhood around the spatial position estimate.  The initial spatial position for each 8-frame chunk is the last spatial position that the point was estimated to be visible.  As PIPs operates only forward in time, we run it twice for each query point: once forward, and once backward in time.  Otherwise, we run the algorithm without modification, directly using its visibility estimates with a threshold of $0.5$ to estimate occlusion.  We found that the chaining algorithm requires more than 30 minutes to process a 250-frame video at $256\times256$ resolution, meaning that we only had time to run on 167 random Kinetics videos before the deadline.  Unlike for COTR, however, these results use all points available in each video. Experiments were run on a GCP machine with 1 V100 GPU and required approximately 100 hours.

\changed{
\section{Semantic analysis}
Although our dataset is not intended for classification, it is useful to consider the distribution of labeled objects to understand performance.  While DAVIS and RGB-stacking are small enough for the object distribution to be observed qualitatively, for Kinetics, we had annotators explicitly label the boxes before adding points.  This process was informal, and we don't control for duplicates or typos; nevertheless, we show the full list of labels as well as their frequencies in Tables~\ref{tab:semantic1}--\ref{tab:semantic4}.  We see a large variety of objects, although unsurprisingly for Kinetics, they tend to be people, clothing, and moving objects that are relevant for human actions, across indoor and outdoor, as well as several types of animals.  We manually broke these into broader categories, and found 26.1\% are on humans, 23.4\% on clothing (on or off the body), and 50.4\% are on other types of objects.  This is consistent with the instructions (to target moving/foreground objects), and we believe it makes the dataset relevant for agents interacting with humans and performing tasks in human environments.  
}
\begin{table*}
\resizebox{\linewidth}{!}{ %
\begin{tabular}{lllll}
person:7012 & flexi:9 & hacksaw:3 & lady:3 & cleaning brush:3 \\
shoe:842 & mini car:9 & statue table:3 & black curtain:3 & sleeveless:3 \\
pant:833 & plant pot:9 & dried palm leaves:3 & electric pencil sharpner:3 & strap:3 \\
tshirt:796 & cat:9 & blue blanket:3 & white jacket:3 & aquarium:3 \\
object:614 & weight plate:9 & blue mat:3 & flowers:3 & green grill:3 \\
shirt:608 & musical instrument:9 & steel tap:3 & window sheet:3 & tumbler:3 \\
hand:542 & clock:9 & silver foil:3 & floor mat:3 & arrow holder:3 \\
car:447 & comb:9 & toy train tracker:3 & decor:3 & roll:3 \\
short:413 & paper roll:9 & gas:3 & frame stand:3 & elastic band:3 \\
t shirt:412 & ac:9 & plastic plate:3 & wash basin:3 & welding machine:3 \\
cap:377 & roller:9 & sledge:3 & bathtub:3 & cotton box:3 \\
chair:372 & lamp light:9 & mini trampoline:3 & showcase plant:3 & steel rods:3 \\
box:370 & screw tighter:9 & pendant:3 & spinning top:3 & white scooter:3 \\
table:347 & garland:9 & snake:3 & hook:3 & propane fueling station:3 \\
bottle:310 & tower:9 & face cream:3 & wall hanger:3 & blue scooter:3 \\
jacket:247 & black coat:9 & headphone:3 & mining helmet:3 & hanging handle:3 \\
helmet:235 & plastic item:9 & violin stick:3 & color box:3 & wall decor:3 \\
ring:223 & barrier:9 & circular frame:3 & sleeve dress:3 & thermocol:3 \\
light:215 & tree pot:9 & skiers:3 & crocodile:3 & green bag:3 \\
watch:214 & crown:9 & system:3 & christmas tree:3 & machine part:3 \\
bowl:209 & chess piece:9 & paint spray machine:3 & white tray:3 & wood object:3 \\
cloth:207 & bedsheet:9 & iron grill:3 & helmate:3 & screwdriver:3 \\
glove:186 & wrist band:9 & marker pen:3 & red shirt:3 & concrete fire bowl:3 \\
machine:186 & pencil:9 & yarn:3 & righthand:3 & side pin:3 \\
bag:176 & glouse:9 & plastic glass:3 & streetlight:3 & whistle:3 \\
pole:167 & yacht:9 & glass bowl:3 & trumpet:3 & icebreaker:3 \\
toy:163 & wall poster:9 & p:3 & whiper:3 & sushi food:3 \\
plate:158 & headphones:9 & pouch:3 & shaving razor:3 & plaster roller:3 \\
spects:150 & red button:9 & mic stand:3 & golfstick:3 & gift wrap:3 \\
ball:147 & blue tshirt:9 & car cover:3 & banner stand:3 & plaster:3 \\
frame:147 & lantern:9 & staff:3 & flex:3 & blue cover:3 \\
wood:140 & black short:9 & chips:3 & bean bag:3 & spects frame:3 \\
door:139 & wood board:9 & handband:3 & egg:3 & snow scooter:3 \\
dress:138 & black top:9 & snow bike:3 & holdingbelt:3 & switch box:3 \\
paper:135 & bat:9 & bed cot:3 & air pump machine:3 & man hole:3 \\
glass:132 & photographer:9 & yellow belt:3 & blue barrel:3 & embroidery hoop ring:3 \\
vehicle:132 & jug:9 & shark:3 & dumbell:3 & staple puller:3 \\
stick:119 & wood block:9 & plastic box:3 & broom stick:3 & cutting plier:3 \\
sofa:117 & wind cart:9 & luggage bag:3 & right gloves:3 & clothes tub:3 \\
coat:115 & electric pole:9 & double tape:3 & plunger:3 & black bucket:3 \\
window:114 & instrument:9 & decoration light:3 & western toilet seat:3 & clothes bag:3 \\
board:108 & light stand:9 & baw:3 & cutting machine:3 & silver stick:3 \\
rod:105 & flying disc:9 & refrigerator:3 & pallet:3 & wooden roller:3 \\
rope:102 & vacuum cleaner:9 & teeth braces:3 & black dress:3 & support board:3 \\
stand:97 & oil bottle:9 & side mirror:3 & blade:3 & metal water pipe:3 \\
tool:93 & mouse:9 & tool box:3 & meat pieces:3 & paper sheet:3 \\
knife:92 & weight lift:9 & machinery equipment:3 & groove machine:3 & automatic door operator:3 \\
flag:92 & watermelon:9 & cable:3 & white short:3 & chef cap:3 \\
pipe:91 & zip:9 & putty plate:3 & measuring tape:3 & green apple:3 \\
bucket:91 & fish:9 & coupling pipe:3 & baby boy:3 & petrol holder:3 \\
band:86 & cleaner:8 & stairs:3 & wooden spoon:3 & car petrol cap:3 \\
mat:84 & card shelf:8 & neck support:3 & white pad:3 & petrol pipe:3 \\
right hand:81 & cushion:8 & glass object:3 & black cup:3 & orange light:3 \\
brush:79 & bandage:8 & gear shifter:3 & red block:3 & white light:3 \\
curtain:76 & red box:8 & drilling machine:3 & shield:3 & red light:3 \\
thread:75 & black t shirt:8 & hinge:3 & electric stove:3 & car bumper:3 \\
hat:75 & red ball:8 & door engis:3 & streeing:3 & dumbbell plate:3 \\
book:75 & wooden board:8 & goal post:3 & metal chrome polish bottle:3 & flex banner:3 \\
left hand:74 & carpet:8 & fedlight:3 & exhaust pipe:3 & podium:3 \\
chain:74 & mixer:7 & yellow bucket:3 & wooden staircase:3 & baby mattress:3 \\
dog:72 & spatula:7 & wood bench:3 & ladder stand:3 & wood tray:3 \\
mirror:72 & fire extinguisher:7 & paino:3 & bubble bottle:3 & sharpner:3 \\
horse:72 & googles:7 & western top:3 & scrubber:3 & tissue box:3 \\
belt:70 & tile:7 & black item:3 & water sprinkler:3 & brown sheet:3 \\
wire:70 & bottle cap:7 & coconut:3 & chandelier:3 & cup board:3 \\
hand band:63 & carry bag:7 & green mat:3 & cot:3 & fruits:3 \\
socket:63 & wall clock:7 & art sheet:3 & pumpkin piece:3 & snow lifter:3 \\
cover:62 & straightener:6 & blue egg:3 & backet:3 & wood cutting machine:3 \\
button:61 & key:6 & pink basket:3 & exercise suit:3 & wood box:3 \\
cycle:60 & elephant statue:6 & yellow egg:3 & watermelon piece:3 & snow toy:3 \\
goggles:58 & specs:6 & dog sofa:3 & torch light:3 & vaccum pipe:3 \\
bike:58 & bridge:6 & left gloves:3 & leather hand band:3 & carry baby bag:3 \\
handle:56 & yoga mat:6 & green button:3 & hoop hula:3 & straw:3 \\
photo frame:55 & mattress:6 & suitcase:3 & rings:3 & pink jacket:3 \\
speaker:55 & plastic bowl:6 & light lamp:3 & white hoody:3 & mitten:3 \\
spectacles:55 & black pant:6 & car engine:3 & soap bubble sticks:3 & pepper crusher:3 \\
\end{tabular}
}
\caption{
\textbf{Full list of object categories named by annotators in \datasetname-Kinetics.} Number of points per category is listed next to the category.
}
\label{tab:semantic1}
\end{table*}
\begin{table*}
\resizebox{\linewidth}{!}{ %
\begin{tabular}{lllll}
basket:54 & green rope:6 & injector pipe:3 & white item:3 & oil container:3 \\
poster:54 & flower:6 & tube:3 & pagdi:3 & mug jar:3 \\
mike:53 & wooden pole:6 & desktop:3 & wooden mud pot:3 & wafer:3 \\
pot:53 & basketball board:6 & support rod:3 & tyrecap:3 & orange rope:3 \\
girl:52 & cutting player:6 & machine tool:3 & black paper:3 & paint bottle:3 \\
mobile:51 & wall:6 & woolen thread:3 & right ring:3 & table cover:3 \\
boy:51 & paper rocket:6 & weaving needle:3 & left bracelet:3 & dolpin:3 \\
switch board:51 & rolling mat:6 & trolly:3 & guitar box:3 & railing bridge:3 \\
packet:51 & doughnut:6 & brown bricks:3 & baby chair:3 & ice box:3 \\
ear ring:48 & washing machine:6 & train toy:3 & refridgerator:3 & cold jar:3 \\
top:47 & clothes:6 & chisel:3 & string:3 & black rod:3 \\
right shoe:46 & motorcycle:6 & orange pencil:3 & food packet:3 & net holder:3 \\
spoon:46 & scooter:6 & white marble:3 & swim glasses:3 & exercise ball:3 \\
bed:46 & left glove:6 & paper punch:3 & traning fin:3 & big basket:3 \\
cup:45 & jack:6 & papers:3 & scuba diving fins:3 & ash shirt:3 \\
tyre:45 & blue jacket:6 & backpack:3 & hookah pipe:3 & wooden base:3 \\
card:45 & red bag:6 & knife box:3 & lorry:3 & hair straightener:3 \\
pan:44 & rim:6 & phone mount:3 & chopping board:3 & blue board:3 \\
equipment:43 & chimney:6 & ornament:3 & cake piece:3 & cement brick:3 \\
apple piece:42 & sandle:6 & motor switch:3 & steel plate:3 & pompoms:3 \\
boat:42 & speedometer:6 & seat cover:3 & decorative:3 & gift box:3 \\
hair band:42 & wind gong:6 & sign:3 & painting board:3 & exam pad:3 \\
jar:42 & nut:6 & milk box:3 & key hole:3 & wooden floor:3 \\
hoodie:41 & number plate:6 & mixer grinder:3 & glass pipe:3 & stretcher:3 \\
net:40 & blue joystick:6 & steel net:3 & frock:3 & powered parachute:3 \\
tray:40 & swim suit:6 & white bricks:3 & music instrument plate:3 & ring box:3 \\
guitar:40 & stage:6 & blue button:3 & crash cymbal:3 & stopper cone:3 \\
left shoe:40 & tire:6 & white machine:3 & card door lock:3 & gas stove:3 \\
stool:39 & digging tool:6 & glass bottle:3 & selfie stick:3 & wheel barrow:3 \\
kid:39 & ribbon:6 & carrot piece:3 & banana:3 & wax strip:3 \\
wheel:39 & steel table:6 & wing screw:3 & basketball hoop:3 & router box:3 \\
bracelet:39 & animal:6 & microphone stand:3 & filmy equipment:3 & small tin:3 \\
parachute:38 & binding rod:6 & sand tray:3 & wooden pad:3 & pipe slide:3 \\
mic:37 & steel rod:6 & brown pant:3 & hand towel:3 & ear stud:3 \\
scissor:36 & necklace:6 & extinguisher:3 & water jug:3 & shorts:3 \\
stone:36 & rightshoe:6 & coke bottle:3 & plates:3 & barbell rod:3 \\
socks:35 & hipbelt:6 & purse:3 & mud remover:3 & score card:3 \\
pillow:35 & display:6 & tripod:3 & leaves:3 & baseball bat:3 \\
television:35 & inner wear:6 & ridge gourd:3 & swim fin:3 & id card:3 \\
rock:35 & log:6 & costume:3 & paint spray:3 & black grill:3 \\
slipper:35 & metal sheet:6 & ridge gourd piece:3 & chain locket:3 & music instrument:3 \\
camera:34 & utensil:6 & pink shirt:3 & ceiling ac vents:3 & pad plates:3 \\
apple:34 & cucumber:6 & cable roller:3 & pink object:3 & pool bridge:3 \\
pen:33 & wall frame:6 & airtrack:3 & mike stand:3 & air meter:3 \\
monitor:33 & goal:6 & nose ring:3 & baby crib:3 & swimming item:3 \\
child:33 & diving shoe:6 & steel tray:3 & wooden stool:3 & eyebrow pencil:3 \\
house:33 & duster:6 & igloo house:3 & metal wall design:3 & eye lash shaper:3 \\
bolt:33 & time adjuster:6 & wooden table:3 & brake:3 & curling machine:3 \\
tub:33 & white bin:6 & vechile:3 & rubber:3 & kids play slide:3 \\
towel:33 & bus:6 & blue short:3 & pad stand:3 & plastic ring:3 \\
black object:33 & wooden cupboard:6 & yellow tshirt:3 & round table:3 & pink tshirt:3 \\
jeans:32 & play card:6 & green short:3 & fruit basket:3 & water skater:3 \\
cupboard:32 & hairclip:6 & woodframe:3 & mike transmitter:3 & sand machine:3 \\
tie:30 & charger:6 & cock:3 & silver nut:3 & extension socket:3 \\
statue:30 & swim cap:6 & mannequin:3 & water pot:3 & paint tube:3 \\
baby:30 & trolley:6 & soft boat:3 & yellow bowl:3 & cotton cake box:3 \\
bangle:30 & clay:6 & fire stand:3 & snow shovel:3 & weight:3 \\
red object:29 & wooden bar:6 & yellow handle:3 & book rack:3 & round comb:3 \\
lamp:29 & blue jeans:6 & stroller:3 & white pipe:3 & creddle:3 \\
white box:28 & signboard:6 & back wheel:3 & black stand:3 & glass plate:3 \\
sheet:28 & bread piece:6 & shell:3 & fire wood:3 & water safety:3 \\
hammer:28 & brown object:6 & beanie:3 & small girl:3 & cake plate:3 \\
dustbin:27 & white cap:6 & yellow rope:3 & chess board:3 & white stool:3 \\
bicycle:27 & dumble:6 & iron handler:3 & pink box:3 & hair roller:3 \\
laptop:27 & wooden chair:6 & mushroom:3 & pink book:3 & dish wash bottle:3 \\
holder:27 & potted plant:6 & strings:3 & meat:3 & solar board:3 \\
bench:27 & head cap:6 & wick:3 & helicopter:3 & screw holder:3 \\
candle:27 & red thread:6 & black board:3 & statue head:3 & nail polish:3 \\
needle:27 & measuring tool:6 & black half sleeve shirt:3 & finger cap:3 & paper packet:3 \\
\end{tabular}
}
\caption{
\textbf{Full list of object categories named by annotators in \datasetname-Kinetics, continued.}
}
\end{table*}
\begin{table*}
\resizebox{\linewidth}{!}{ %
\begin{tabular}{lllll}
ladder:26 & shoe rack:6 & oxygen tank:3 & steel item:3 & peeler:3 \\
phone:25 & white block:6 & toy track:3 & girl dress:3 & oil tub:3 \\
sign board:25 & glass jar:6 & purple jacket:3 & short pant:3 & locker:3 \\
tree:25 & food:6 & fuel can:3 & surfboard:3 & water can:3 \\
dumbbell:25 & painting:6 & match stick:3 & drum plate:3 & support stand:3 \\
rack:25 & baby bed:6 & pink top:3 & door closer:3 & white band:3 \\
bow:24 & orange:6 & footwear:3 & wood plank:3 & dust bin:3 \\
photo:24 & orange slice:6 & stove stand:3 & gear:3 & pine apple:3 \\
mascara:24 & giraffe:6 & white flower:3 & plastic knob:3 & white blind:2 \\
doll:24 & sheep:6 & tape dispenser:3 & wooden rack:3 & right sock:2 \\
fan:24 & metal candle cup:6 & balloon toy:3 & snowman:3 & sketch:2 \\
tap:24 & glass table:6 & ballon:3 & wind instrument:3 & pink thread:2 \\
gloves:24 & tomato:6 & supporter:3 & safety cap:3 & filter:2 \\
cutter:24 & steel object:6 & aeroplane:3 & window glass:3 & saddle:2 \\
earring:24 & stainer:6 & pineapple:3 & dash board:3 & stethoscope:2 \\
sweater:23 & projector:6 & pink bin:3 & electric engraving machine:3 & foam:2 \\
cake:21 & steel box:6 & cards:3 & wood stand:3 & white cardboard:2 \\
blue cloth:21 & white strip:6 & right indicator:3 & orangestool:3 & hinges:2 \\
spray bottle:21 & sponge:6 & left indicator:3 & calender:3 & saddle rope:2 \\
dolphin:21 & door handle:6 & match box:3 & blinds:3 & news paper:2 \\
sticker:21 & blue top:6 & broom:3 & vaccum machine:3 & doormate:2 \\
balloon:21 & head scarf:6 & roti maker:3 & game equipment:3 & bull ride:2 \\
fence:21 & cradle:6 & cleaning machine:3 & game board:3 & pineapple slice:2 \\
drum:21 & light pole:6 & cabin:3 & coolent box:3 & box cap:2 \\
steering:21 & head band:6 & card board:3 & funnel:3 & grinder:2 \\
banner:21 & boxing bag:6 & food box:3 & infuser:3 & bracelette:2 \\
fork:21 & cd box:6 & pricetag:3 & bike delivery box:3 & wind manipulator:2 \\
block:21 & wardrobe:6 & stamps:3 & carrot:3 & underwear:2 \\
screw:21 & door lock:6 & gum tube:3 & wooden pot:3 & white napkin:2 \\
tin:21 & woman:6 & number sheet:3 & showpiece:3 & robot toy:2 \\
black box:21 & rubber band:6 & roughfile:3 & stainless steel water jug:3 & mobile holder:2 \\
yellow object:20 & dish:6 & label:3 & cigarette pipe:3 & oxygen pumping machine:2 \\
berry:20 & piano:6 & softdrink tin:3 & bluetooth:3 & gym bench:2 \\
fridge:20 & train:6 & arrow:3 & watering pipe:3 & tennis bat:2 \\
white board:20 & scale:6 & foot:3 & radio:3 & red rod:2 \\
blue shirt:20 & gear rod:6 & yellow short:3 & decor light:3 & green pant:2 \\
stove:20 & slide:6 & dressing table:3 & disco light:3 & orange shirt:2 \\
remote:20 & water bottle:6 & cristmas tree:3 & music box:3 & violet dress:2 \\
oven:19 & mixer jar:6 & jumping castle:3 & drum stick:3 & left handle:2 \\
napkin:19 & yellow thread:6 & machine top:3 & music plate:3 & flower balloons:2 \\
hanger:18 & machine rod:6 & bike mirror:3 & bell plate:3 & yellow button:2 \\
potato:18 & wool:6 & front shield:3 & windgong:3 & frontwheel:2 \\
steel:18 & battery box:6 & swim board:3 & air pump:3 & black band:2 \\
white object:18 & microphone:6 & torch:3 & spading stick:3 & watertin:2 \\
vest:18 & iron sheet:6 & bra pad:3 & pool lane rope:3 & pertson:2 \\
tissue:18 & thread bundle:6 & bra tag:3 & bee box:3 & vehcile:2 \\
cow:18 & black button:6 & sewing machine:3 & steel glass:3 & hanging light:2 \\
skate board:18 & car seat:6 & blue object:3 & tea pot:3 & fishing stick:2 \\
lighter:18 & white t shirt:6 & cotton:3 & sleeve shirt:3 & plane:2 \\
flute:18 & crane:6 & cleaning mop:3 & fencing mesh:3 & fishing rod:2 \\
hand glove:18 & weight equipment:6 & decoration item:3 & cooling glasses:3 & plastic disk:2 \\
black bag:18 & spray:6 & white ball:3 & sticky notes:3 & iron stand:2 \\
shovel:18 & red carpet:6 & white boat:3 & album:3 & table fan:2 \\
screen:18 & steamer:6 & legs:3 & stick paper:3 & shed:2 \\
apron:18 & wooden bench:6 & wallet:3 & file handle:3 & goat:2 \\
headband:17 & commode:6 & globe:3 & hand kerchief:3 & head cover:2 \\
plant:17 & wristband:6 & trimble scanning:3 & id tag:3 & fencing rod:2 \\
seat:17 & wiper:6 & avacado:3 & snow stick:3 & dartboard:2 \\
sock:16 & sword:6 & handles:3 & metal tool:3 & blue banner:2 \\
red cloth:16 & leftshoe:6 & rubiks cube:3 & treadmill:3 & textperson:2 \\
knob:16 & scissors:6 & right mirror:3 & spin bike:3 & shampoo bottle:2 \\
coin:15 & books:6 & metal wire:3 & left slipper:3 & tperson:1 \\
ship:15 & green box:6 & chopstick:3 & tongs:3 & boxcap:1 \\
umbrella:15 & extension box:6 & nail cutter:3 & wrist belt:3 & parchute:1 \\
blanket:15 & soap:6 & hatchet:3 & handwash:3 & bracelettle:1 \\
blue box:15 & right leg:6 & black file:3 & honey tray:3 & letter e:1 \\
pin:15 & traffic cone:6 & hand catcher:3 & shutter gear box:3 & dishwasher:1 \\
tv:15 & walkie talkie:6 & speed meter:3 & gear box:3 & frige:1 \\
mask:15 & black stick:6 & inhaler:3 & shop:3 & shie:1 \\
drill machine:15 & ski:6 & horn switch:3 & football net:3 & wodd:1 \\
brick:15 & trunk belt:6 & letter m:3 & sink:3 & coffee maker:1 \\
cabinet:15 & boot:6 & t shirt roll:3 & breather:3 & blue suit:1 \\
tape:15 & tab:6 & menu book:3 & handle knob:3 & pine apple slice:1 \\
cylinder:15 & weight lifting equipment:6 & green jacket:3 & red machine:3 & bandage clip:1 \\
camel:15 & blue pant:6 & ski helmet:3 & show case:3 & new paper:1 \\
windmill:15 & watering can:6 & traffic light:3 & yellow stone:3 & pany:1 \\
red cap:15 & left leg:6 & steel bowl:3 & chappal:3 & newspaper:1 \\
leg:15 & disk:6 & water melon piece:3 & white background:3 & water melon peice:1 \\
gate:15 & weight lifting:6 & toy brick:3 & wrapper:3 & show peice:1 \\
bird:15 & cigar pipe:6 & flip flop:3 & land line:3 & dart board:1 \\
cardboard:15 & lefthand:6 & left socks:3 & ventilator:3 & white cupboard:1 \\
\end{tabular}
}
\caption{
\textbf{Full list of object categories named by annotators in \datasetname-Kinetics, continued.}
}
\end{table*}
\begin{table*}
\resizebox{\linewidth}{!}{ %
\begin{tabular}{lllll}
hand bag:15 & trimmer:6 & knife holder:3 & suit:3 & head:1 \\
van:15 & mortar board holder:6 & megaphone speaker:3 & sleeveless tshirt:3 & oxygen pumpimg machine:1 \\
wooden box:15 & metal:6 & barricating floor stand:3 & detector:3 & wall poster top:1 \\
iron rod:15 & wrench:6 & porch swing:3 & sniper box:3 & cushoin:1 \\
pad:15 & spanner:6 & watertank:3 & thermocol sheet:3 & galss:1 \\
axe:15 & basin:6 & tiger:3 & track pant:3 & frame top:1 \\
berries:15 & couch:6 & black suit:3 & white bucket:3 & peerson:1 \\
wooden object:15 & bud:6 & wooden shelf:3 & green truck:3 & pizza oven:1 \\
gun:15 & white tshirt:6 & red clip:3 & circle balloon:3 & gyn bench:1 \\
lock:15 & chips packet:6 & hairpin:3 & toy block:3 & branch:1 \\
tent:15 & burner:6 & mobile case:3 & traffic light toy:3 & preson:1 \\
idol:15 & vase:6 & wood blocks:3 & spectacle:3 & sipper:1 \\
can:14 & eyelash curler:6 & black pipe:3 & baby dress:3 & street light:1 \\
hockey stick:14 & marble:6 & blue block:3 & baby apron:3 & headlight:1 \\
scarf:14 & ribbon roll:5 & finger tool:3 & sausage:3 & cardshelf:1 \\
cone:14 & blue sheet:5 & green shirt:3 & blue chalk:3 & persoon:1 \\
black shirt:14 & container:5 & glue gun tool:3 & soap stand:3 & greenpipe:1 \\
white cloth:14 & pig:5 & iron frame:3 & gamepad:3 & glasse middle:1 \\
clip:13 & chip:5 & umberlla stand:3 & yellow toy:3 & glass right:1 \\
white sheet:13 & guage:5 & bathing basin:3 & door handler:3 & glass left:1 \\
flower vase:13 & tube light:5 & plastic curtain:3 & purple bottle:3 & laddder:1 \\
bin:12 & left mirror:5 & fan hanger:3 & c c camera:3 & vihicle:1 \\
grill:12 & kite:5 & rod piece:3 & hairband:3 & paper shredder:1 \\
bed sheet:12 & playing machine:5 & orange tool:3 & hut:3 & frame middle:1 \\
black jacket:12 & dish washer:5 & maroon cloth:3 & yellow paper:3 & persson:1 \\
gym equipment:12 & ice axe:5 & yellow cloth:3 & syringe:3 & form:1 \\
file:12 & photoframe:5 & telephone:3 & metal piece:3 & tbowl:1 \\
duck:12 & show piece:5 & car number plate:3 & drawer:3 & boardboard:1 \\
headset:12 & wooden log:5 & blue thread:3 & toy house:3 & plastic desk:1 \\
pumpkin:12 & skipping rope:5 & toy stick:3 & marker:3 & whiteshirt:1 \\
black cloth:12 & white shirt:5 & olive oil bottle:3 & barricade:3 & palne:1 \\
sandal:12 & hoddie:5 & detecting machine:3 & hand break:3 & bowling equipment:1 \\
sprayer:12 & cpu:5 & mixing bowl:3 & wooden piece:3 & hooddie:1 \\
puzzle:12 & dining table:5 & steel tool:3 & tambourine:3 & res cloth:1 \\
elephant:12 & point:5 & paint filler:3 & guitar bag:3 & red scarf:1 \\
chess coin:12 & track:5 & piano keyboard:3 & white bottle:3 & yelllow object:1 \\
tissue roll:12 & hair dryer:4 & tv stand:3 & glass door:3 & tabel:1 \\
skirt:12 & gauge:4 & toy car:3 & stop indicator board:3 & plastic object:1 \\
wooden block:12 & building:4 & cutting board:3 & leaning bench:3 & mug:1 \\
tong:12 & black thread:4 & sound box:3 & bar lifting:3 & balack band:1 \\
wood piece:12 & motor:4 & tambourine hand percussion:3 & threadmill:3 & front wheel:1 \\
wooden plank:12 & iron box:4 & pool lane divider:3 & mosquito killer:3 & spects left:1 \\
diaper:12 & power box:4 & desk:3 & shelf:3 & white rolling button:1 \\
wrist watch:12 & wooden stick:4 & wooden tray:3 & dvd player:3 & person back shoulder:1 \\
racket:12 & blue funnel:3 & bread:3 & paper crusher:3 & person forehead:1 \\
lid:12 & handler:3 & red jacket:3 & auto:3 & cupboard top:1 \\
flower pot:12 & plucker tool:3 & work piece:3 & woollen:3 & cupboard middle:1 \\
glasses:12 & skate boarding:3 & red blanket:3 & blue color object:3 & cupboard bottom:1 \\
hair clip:12 & iron lid:3 & cloths:3 & green object:3 & personn:1 \\
carton box:12 & greentub:3 & pebble:3 & person hand:3 & rind:1 \\
truck:12 & cement lid:3 & trowel:3 & candle cap:3 & door mate:1 \\
joystick:12 & router:3 & red suit:3 & cube:3 & jeans jacket:1 \\
keyboard:12 & yellow carpet:3 & rope ball:3 & brown box:3 & white card board:1 \\
bulb:12 & tractor:3 & woolen roll:3 & paper glass:3 & ttshirt:1 \\
screw driver:12 & baby bottle:3 & man:3 & wood machine:3 & table right:1 \\
stud:12 & kids walker:3 & talcum powder:3 & light box:3 & table middle:1 \\
plastic cover:12 & store:3 & drink bottle:3 & laptop table:3 & table left:1 \\
right glove:11 & dumbbell set:3 & turban:3 & rack cupboard:3 & parachute skydiving:1 \\
fruit:11 & left ear stud:3 & ash tshirt:3 & tyer:3 & nuts packet bottom:1 \\
switch:11 & needle bag:3 & round disk:3 & orange tshirt:3 & perason:1 \\
green pipe:11 & parachute rope:3 & safety jacket:3 & rod stand:3 & blue stone:1 \\
text:11 & controller:3 & plank:3 & red curtain:3 & swimming goggles:1 \\
switchboard:10 & violin:3 & teddy:3 & dryer:3 & hesd cover:1 \\
kangaroo:10 & cellphone:3 & chart:3 & railing:3 & photofrane:1 \\
plastic bag:10 & kids slide:3 & cake pan:3 & topwear:3 & balll:1 \\
paint brush:9 & red toy:3 & orange object:3 & strainer:3 & nuts packet top:1 \\
egg yolk:9 & electrical box cover:3 & tank:3 & brown bag:3 & nuts packet middle:1 \\
head light:9 & big spoon:3 & room cleaner:3 & mascara stick:3 & frame bottom:1 \\
black tshirt:9 & left sock:3 & wheel cap:3 & pink sheet:3 \\
\end{tabular}
}
\caption{
\textbf{Full list of object categories named by annotators in \datasetname-Kinetics, continued.}
}
\label{tab:semantic4}
\end{table*}

\end{document}